\documentclass[11pt,letterpaper]{article}

\usepackage[T1]{fontenc}
\usepackage[utf8]{inputenc}
\usepackage{newtxtext,newtxmath}
\usepackage[letterpaper,margin=1in]{geometry}
\usepackage{microtype}
\usepackage{graphicx}
\usepackage{booktabs}
\usepackage{amsmath}
\usepackage{algorithm}
\usepackage{algorithmic}
\usepackage{caption}
\usepackage{subcaption}
\usepackage{adjustbox}
\usepackage[authoryear,round]{natbib}
\usepackage[hyphens]{url}
\usepackage{xcolor}
\usepackage{hyperref}

\IfFileExists{paper_body.tex}{

  \newcommand{\bibliographydatabase}{sharding}
  \graphicspath{{./}}
}{

  \newcommand{\bibliographydatabase}{arxiv/sharding}
  \graphicspath{{arxiv/}}
}

\definecolor{linkblue}{HTML}{244F73}
\hypersetup{
  colorlinks=true,
  linkcolor=linkblue,
  citecolor=linkblue,
  urlcolor=linkblue,
  pdfborder={0 0 0},
  pdftitle={Sharding Prevents LLM Oversight Failures and Adversarial Exploitation},
  pdfauthor={Victor Akinwande, J. Zico Kolter, Aran Nayebi}
}
\title{Sharding Prevents LLM Oversight Failures\\
  and Adversarial Exploitation}
\author{
  Victor Akinwande\textsuperscript{1}%
  \thanks{Correspondence to: \texttt{vakinwan@andrew.cmu.edu}.}
  \quad
  J. Zico Kolter\textsuperscript{1}
  \quad
  Aran Nayebi\textsuperscript{1,2}
  \\[0.55em]
  \small
  \textsuperscript{1}School of Computer Science
  \quad
  \textsuperscript{2}Neuroscience Institute
  \\[0.3em]
  \normalsize Carnegie Mellon University
  \\[0.3em]
  \small Preprint
}
\date{}

\begin{document}

\maketitle

\begin{abstract}
Giving an LLM judge more compute does not necessarily make it check more requirements. When one call must return many verdicts, some decisions become weakly grounded in the evidence, even when that call receives the same token or tool budget as a panel of separate calls. Across expert-graded research replications, legal work, and clinical-trial assessments, agreement with experts falls as the number of verdicts per call grows. We identify \emph{sharding} as the intervention that mitigates this failure in model-based oversight. Sharding partitions the requirements into smaller groups, assigns each group to a separate call, and aggregates the verdicts. Against a single call with the panel's full budget, sharding improves agreement while holding the model, evidence, total budget, and per-decision budget fixed. \emph{Overall, we find that a sharded weaker judge can outperform a more capable holistic judge and match that judge even when the latter receives the panel's full budget}.

Additionally, we find that sharding exhibits robustness against adversaries.
A best-of-$N$ adversary can hold the underlying work fixed, vary only its presentation, and increase an overloaded judge's acceptance of genuinely unmet criteria severalfold.
Wherever sharding reduces baseline error, it removes this adversarial advantage, keeping over-acceptance low even as the adversary's search widens. Sharding does not address attacks that persuade the judge separately on each criterion rather than exploiting overload. In that setting, we find that debate-style opposition on top of sharding withstands such adaptive re-optimization.
\end{abstract}

\section{Introduction}
\label{sec:intro}

Language models produce work faster and at larger scale than people can check by hand, and the checking is increasingly delegated to other models: a model verifies code against held-out tests, judges a long response against a detailed rubric, or serves as a judge in an evaluation pipeline \citep{zheng2023judging,bowman2022measuring}. The stakes are highest when the work under review may contain failures its producer is motivated to present as success, such as a reward-hacked solution that passes the visible tests while violating the intent \citep{pan2022effects,gao2023scaling}.
In that setting, the oversight model performs adversarial detection, and a model that misses such a failure has been gamed. A common design uses one capable model, presenting it with the artifact and the full set of requirements at once, so that a single submission may carry tens or hundreds of independent checks graded in one call.

We test three empirical claims. First, simply giving one judge more computation does not fix the problem of asking it to assess too many requirements at once. Sharding improves accuracy by assigning smaller groups of requirements to separate judge runs, even with the same total computational budget. We demonstrate this on expert-labeled evaluations of research replications, legal work, and clinical trials. Second, these holistic judge errors are exploitable: a producer can present the same work in several ways and select the version most likely to make the judge accept unmet requirements. Our best-of-$N$ presentation attack succeeds against both graders and code verifiers. Third, sharding limits this attack when overload causes the errors, while combining sharding with within-shard methods such as debate provides greater robustness to adaptive attacks.

We make five contributions. \textbf{(i)} We show that decision load degrades accuracy and that sharding recovers accuracy that additional budget in one holistic judgment does not, using real expert grades from three independent datasets and domains: research-replication grading (PaperBench), legal-work grading (JudgmentBench), and clinical-trial risk-of-bias assessment (ROBoto2) (Section~\ref{sec:accuracy}). \textbf{(ii)} We instantiate a best-of-$N$ presentation attack that holds ground truth fixed and exploits this overload, driving an overloaded judge's over-acceptance severalfold and raising a code verifier's cleared-defect rate from 0.12 to 0.44 (Section~\ref{sec:attack}). \textbf{(iii)} We show that sharding closes this attack surface by ensuring that each criterion receives scrutiny, reducing over-acceptance on all four graded submissions. The defense is strongest when decisions compete for attention, but does not stop attacks that persuade the judge on each criterion individually, which require an attack-matched structure (Section~\ref{sec:defense}). \textbf{(iv)} We search the space of within-shard communication structures (opposition, reconciliation, cross-examination, within-criterion decomposition) and find all of them second-order relative to the division itself (Section~\ref{sec:withinshard}). \textbf{(v)} We give a capacity interpretation in which per-decision attention factorizes into budget and load terms, explaining when and at what group size division helps (Section~\ref{sec:interp}).

\section{Setup}
\label{sec:setup}

\subsection{Unit of Analysis and the Oversight Arms}

The unit of analysis is a single decision: whether a specific component of a code artifact, or a specific criterion in a rubric, is satisfied. Each decision carries a ground-truth label the oversight model never sees; an unsatisfied decision is a genuine failure, and clearing it is the error that oversight must prevent. For an artifact with $K$ decisions indexed $k \in \{1,\dots,K\}$ and hidden labels $y_k \in \{0,1\}$ (with $y_k=1$ a failure), an oversight configuration returns scores $s \in [0,1]^K$. The atom of every configuration is one call $\mathrm{call}(D,E,b)$, given the full artifact evidence $E$, a set $D$ of decisions to return, and a budget $b$, producing $\{s_k : k \in D\}$; we write $n=|D|$ for its decision load. The arms hold the oversight model fixed and vary only the allocation of the $K$ decisions to calls and the budget per call. Let $B$ denote the base per-call budget and $S$ the number of shards.

\begin{itemize}
\item \textbf{Solo} (holistic, in grading): one call renders all $K$ decisions at budget $B$; this is the cheapest baseline.
\item \textbf{Full-budget solo}: one call renders all $K$ decisions with the panel's full combined budget, namely $S$ times $B$ tool calls for verification, or a token budget scaled with $K$ up to a cap with a step-by-step instruction for grading; this control isolates division of work from total compute.
\item \textbf{Pooled opinions}: $S$ calls each render all $K$ decisions at budget $B$, aggregated per decision. Matches the sharded panel in total budget but duplicates rather than divides.
\item \textbf{Sharded}: the $K$ decisions are partitioned into $S$ disjoint groups, each judged by one call at budget $B$, so the model attends to a fraction of the decisions per call.
\end{itemize}

The budget accounting isolates decision load. Sharded, pooled opinions, and full-budget solo spend the same total budget $SB$. Sharded and full-budget solo additionally spend the same per-decision budget $SB/K$, and differ only in the decision load per call, $n=K$ for full-budget solo against $n=K/S$ for sharded. The contrast (sharded minus full-budget solo) therefore varies decision load alone at fixed total and per-decision budget. Algorithm~\ref{alg:shard} (Appendix~A) states the sharding procedure, which divides the decisions while leaving the evidence and the per-call budget intact.

\subsection{Metrics, Evidence Standard, and Datasets}

We report AUC where labels and scores are calibrated (code verification) and Cohen's kappa \citep{cohen1960kappa} against human labels where the positive rate varies (grading), with bootstrap 95 percent confidence intervals resampling items. Contrasts are reported under two inferential units where they differ: the decision level, which is sensitive but anticonservative when decisions share context, and the item or task level (a cluster resample), which is conservative. Throughout, a contrast is reported as a main finding only when it survives the conservative unit; effects significant at the decision level alone are labeled suggestive. For the security results we report the over-acceptance rate, the fraction of genuinely failed decisions that the model accepts as satisfied; the defense is the reduction in over-acceptance that sharding produces.

We use three grading datasets with expert labels in three domains: the JudgeEval component of PaperBench \citep{starace2025paperbench} (expert-graded research-replication submissions with one hundred to two thousand leaf criteria per submission), JudgmentBench \citep{yang2026judgmentbench} (expert-lawyer grading of legal work products against professional rubrics; Appendix~A), and ROBoto2 \citep{hevia2025roboto2} (expert Cochrane RoB2 \citep{sterne2019rob2} risk-of-bias assessments of clinical trials; Appendix~A). We also use SWE-Bench Pro \citep{swebenchpro2025} to test a direct-reading case, an oracle-anchored cyber-verification dataset built from confirmed ARVO memory-safety duals \citep{mei2024arvo}, a controlled synthetic grading dataset, and FActScore \citep{min2023factscore} as a structural control. We use ``verifier'' for the code setting and ``judge'' for the grading setting; when a statement covers both, we write ``oversight model.''

\section{Sharding Reduces Decision-Load Errors; More Single-Call Budget Does Not}
\label{sec:accuracy}

\begin{figure}[t]
\centering
\includegraphics[width=0.95\textwidth]{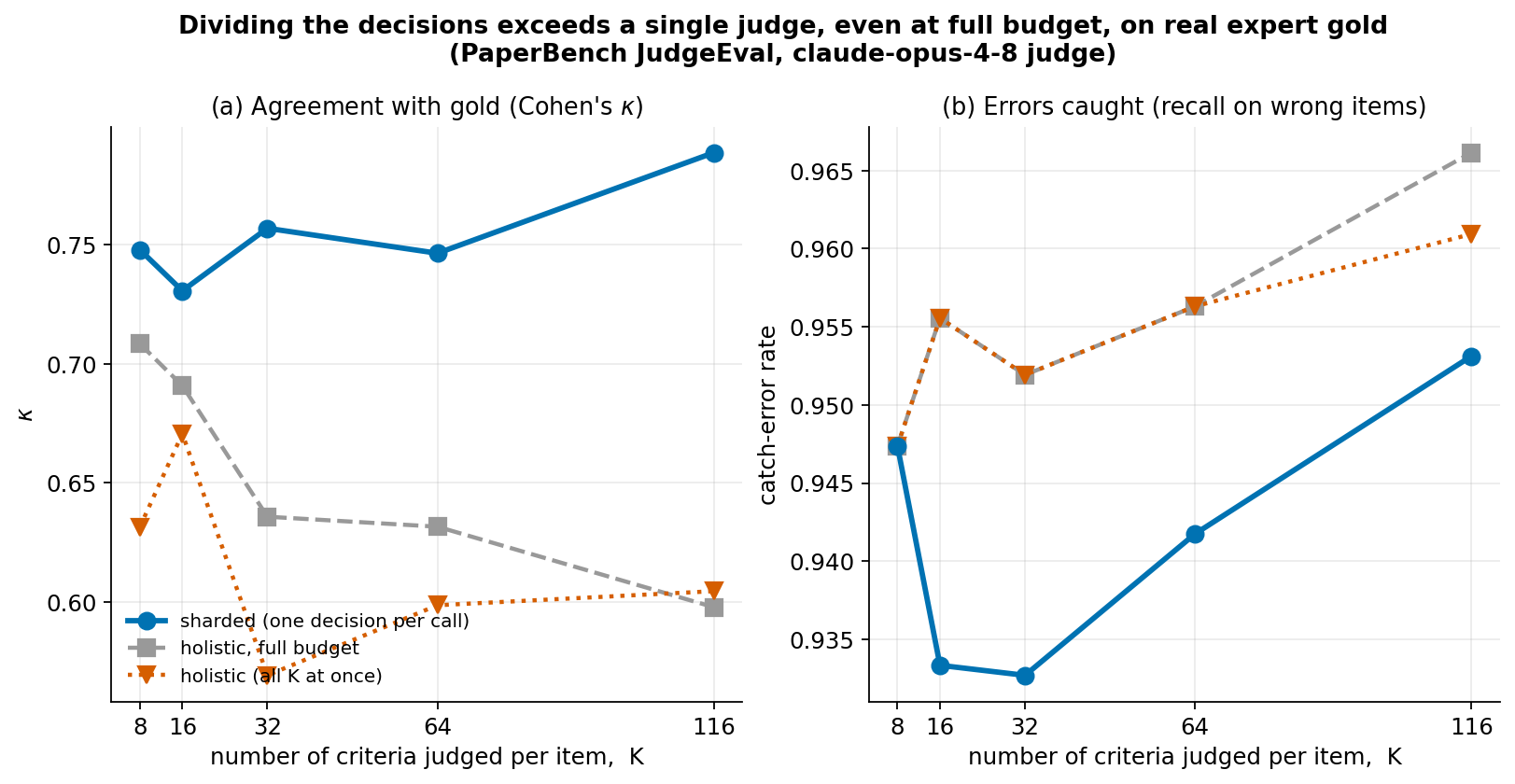}
\caption{On the expert-graded replication dataset with the full submission held as context, agreement against decision load $K$ for each arm (judge Claude Opus 4.8). Sharded grading holds between 0.73 and 0.79 at every load while the single judge and the single judge at full budget sit near 0.60, and the full-budget single judge does not recover, indicating a decision-attention bottleneck. The same construction at the Haiku, Sonnet, and Opus 4.6 tiers is in Appendix~B (Table~\ref{tab:capsweep}).}
\label{fig:loadcurve}
\end{figure}

Using real expert grades from PaperBench, we find that division improves agreement when more single-call budget does not. Fixing one PaperBench submission as complete context and sweeping the number of leaf criteria per judge call, the strongest sharded judge tested (Claude Opus 4.8) holds between kappa 0.73 and 0.79 against the expert grades at every load. At the full load of 116 criteria, the sharded judge reaches 0.789, compared with 0.604 for the single judge and 0.598 for the full-budget single judge (Figure~\ref{fig:loadcurve}). Pooled over loads, the contrast (sharded minus full-budget judge) is 0.142 (CI 0.107 to 0.178), and the margin grows with load (Table~\ref{tab:opusload}). The full-budget single judge does not improve, which locates the bottleneck in decision attention, not total compute. The result holds down the capability range on the same construction, with (sharded minus full-budget judge) of 0.076, 0.096, and 0.040 for Haiku 4.5, Sonnet 4.6, and Opus 4.6, all with intervals excluding zero (Table~\ref{tab:capsweep}), and on a second submission (0.083, CI 0.044 to 0.117, Sonnet 4.6; the full-load sweep on that submission reaches +0.140 at 174 criteria, Appendix~\ref{app:robustness}). Division also substitutes for judge capability: the sharded Sonnet 4.6 judge (kappa 0.75) exceeds the stronger Opus 4.6 judge reviewing holistically (0.68) and matches it at full budget (0.75; Table~\ref{tab:capsweep}), and on the clinical dataset below, the divided weaker judge closes over half the gap between its own holistic agreement and the stronger judge's. Across the tested Claude tiers, the tolerable group size grows with judge capability, and the strongest judge shows no decline on the clinical workload (Tables~\ref{tab:capsweep}, \ref{tab:rob}). The practical rule is capacity-sized grouping rather than maximal sharding. Because each call re-sends the context, an intermediate group is far cheaper than one decision per call without sacrificing agreement: a group of thirty-two criteria reaches kappa 0.86, compared with 0.73 at one criterion per call, while using about one twenty-third as many input tokens (Appendix~B).

We also compare sharding with the strongest single-call baseline we could construct. Beyond the full-budget step-by-step control, we scaffold the single judge with a forced per-criterion evidence citation and a structured checklist table in which the judge writes, for each criterion, its identifier, the single supporting quote from the submission, and a verdict before answering. On JudgmentBench at moderate load the table scaffold matches sharding (kappa 0.614 against 0.611, with every output parsed successfully), so a well-structured single call can maintain per-decision attention when the criteria are few. This works only below a load ceiling. Sweeping the leaf criteria per call on the Stay-on-Topic submission with the strongest judge tested (Opus 4.8), the sharded margin over the table scaffold grows monotonically with load, +0.019, +0.056, +0.129, and +0.173 kappa at sixteen, thirty-two, sixty-four, and one hundred sixteen criteria (intervals excluding zero at thirty-two and above). At the highest load the table scaffold falls below plain holistic evaluation (0.570 against 0.619), so the in-call structure itself degrades as the number of criteria grows. Every output is parsed successfully at every load, so output truncation does not explain the gap (Figure~\ref{fig:opusscaffold}). At the Sonnet tier the same pattern appears (Table~\ref{tab:scaffold}). Division is therefore the scalable way to maintain per-decision scrutiny: a structured single call helps, and even matches division at low load, but cannot sustain the structure as the number of decisions grows.

\begin{figure}[t]
\centering
\includegraphics[width=0.95\textwidth]{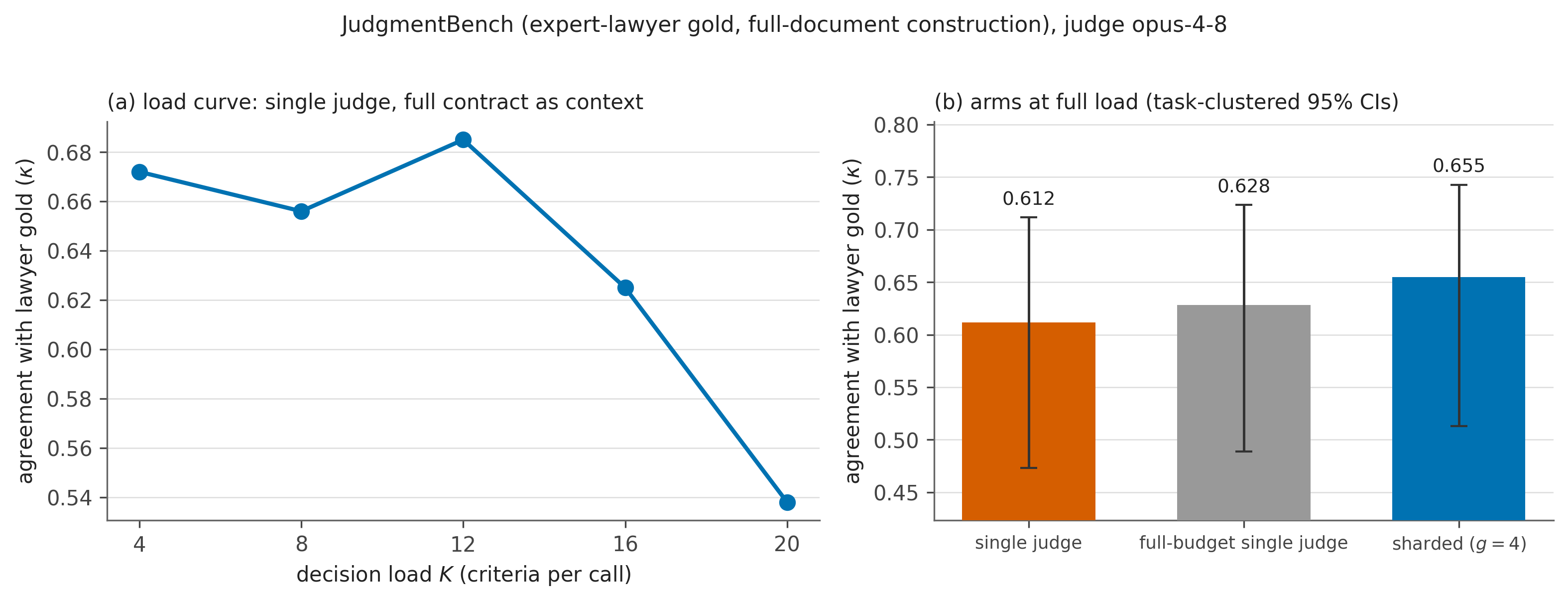}
\caption{Second dataset with expert labels (JudgmentBench, expert-lawyer grading of legal work products, full source contract as context, judge Claude Opus 4.8). (a) A single judge's agreement with the lawyer labels falls as the number of criteria per call grows, with the decline beginning near twenty criteria; decision load therefore limits accuracy even for the strongest model tier tested. (b) At full load a sharded judge exceeds both the single judge and the full-budget single judge, and the full-budget single judge does not close the gap. Error bars are task-clustered 95 percent intervals over the thirteen legal tasks. Sonnet-tier results are in Appendix~B (Tables~\ref{tab:jbacc} and~\ref{tab:jbcap}).}
\label{fig:jbacc}
\end{figure}

The accuracy result replicates on a second dataset with expert labels in a different domain. JudgmentBench supplies expert-lawyer grades of legal work products against professional rubrics. To ensure that each decision requires an integrated model of a large artifact, we prepend the full source contract as shared context, matching the conditions under which the lawyers produced the grades. The contracts have a median length of 34,000 tokens and reach 183,000 tokens. Under this construction a single judge's agreement with the lawyer labels falls as the number of criteria per call grows, and the workload exceeds even the strongest judge's capacity. For Opus 4.8, agreement falls from kappa 0.672 at four criteria to 0.538 at twenty. At full load the sharded judge (group of four) reaches 0.655, compared with 0.612 for the single judge and 0.628 for the full-budget single judge. Both contrasts exclude zero over the thirteen tasks: 0.043 (0.004 to 0.091) and 0.027 (0.002 to 0.052), respectively, under the conservative task-clustered unit (Figure~\ref{fig:jbacc}; Table~\ref{tab:jbcap}). The benefit grows as capability falls, 0.099, 0.063, and 0.043 over the single judge for Haiku 4.5, Sonnet 4.6, and Opus 4.8; on the fuller Sonnet-tier sample the contrast is 0.081 with a task-clustered interval of 0.009 to 0.148, while the Sonnet contrast against the full-budget control, 0.039, excludes zero over decisions but not over tasks and is reported as suggestive (Tables~\ref{tab:jbacc}, \ref{tab:jbcap}).
ROBoto2 tests the result in a third domain. It provides fully manual expert Cochrane risk-of-bias assessments of clinical-trial reports: each of 187 trials carries the 22 standardized RoB2 signaling questions, integrative methodological judgments aggregated by a fixed algorithm to an overall risk verdict. The 22 questions are demanding but few, so whether they overload a judge depends on the judge. They overload the weaker judge (Haiku 4.5), for which dividing the questions improves agreement with the expert answers monotonically as the group shrinks, from the holistic 0.340 to 0.416 at one question per call. This gain of 0.076 (interval 0.033 to 0.117 over trials) excludes zero at group sizes of one and two, while the full-budget single judge does not move from 0.340. The stronger judges are flat at every group size because the 22 questions do not exceed their capacity (Sonnet 4.6 at 0.435 to 0.467 against holistic 0.470, Opus 4.8 at 0.372 to 0.422 against 0.405; Table~\ref{tab:rob}; Figure~\ref{fig:rob}). This dataset demonstrates the division result on expert clinical labels and shows directly that the load at which division helps scales with judge capability.

\textbf{Scope conditions.} Division helps when a single model's accuracy declines as the number of decisions per call grows and additional budget does not reverse the decline. It has little or no benefit otherwise. When each component can be checked by direct reading, a single verifier handles even 64 bundled patches without degrading, and division has a null to slightly negative effect (SWE-Bench Pro, Table~\ref{tab:swebp}). When insufficient budget rather than decision load limits accuracy, the full-budget control recovers completely and division is unnecessary (the synthetic arithmetic dataset). FActScore provides a structural control: despite many decisions over shared context, its individually easy fact checks show no accuracy decline at any tested load (Table~\ref{tab:factscore}). Section~\ref{sec:interp} interprets these conditions jointly.

\section{Non-Sharded Judges Are Adversarially Exploitable}
\label{sec:attack}

The single model's failure under load is not random error but a directional, selectable bias. A forced-evidence audit that requires each call to report the evidence it used and whether it checked each decision shows the mechanism directly (Table~\ref{tab:audit}): the overloaded judge over-flags met criteria (over-flag rate 0.74 to 0.88, near-zero miss) and reports leaving 0.73 to 0.84 of its decisions unchecked, so a fraction of its decisions are settled without support.

We show this blind spot is exploitable with a best-of-$N$ presentation attack. The attack holds ground truth fixed and varies only the presentation of the artifact: for grading, an author-supplied report that argues every criterion is met; for code, a pull-request description that argues the patch is correct. The adversary generates $N$ presentations, scores each with the oversight model, and selects the one that maximizes acceptance. Because ground truth is held fixed, any unmet decision that the selected presentation causes the model to accept is a real over-acceptance. Against the overloaded holistic judge on PaperBench the attack succeeds at both judge tiers tested: with judge and adversary Opus 4.8, best-of-8 selection at least doubles the over-acceptance rate on expert-unmet criteria on every submission, from 0.152 to 0.408 on all-in-one and from 0.080 to 0.303 on the pinn templates (Table~\ref{tab:opusattack}); at the Sonnet tier it drives the rate from about 0.1 at a single presentation to about 0.4 on the most vulnerable submissions (all-in-one and rice), increasing monotonically with $N$ and remaining positive on all four graded submissions (Table~\ref{tab:attackgrading}). Against a code verifier reviewing model-generated patches, best-of-8 selection over pull-request descriptions raises the fraction of defective patches cleared from 0.12 to 0.44 and lowers the verifier's discrimination from AUC 0.71 to 0.56; a single honest description does not help the adversary and in fact raises discrimination to 0.88, so the exploit is a property of the selection over presentations rather than of any one presentation (Figure~\ref{fig:codeattack}). On the oracle cyber-verification dataset the attack raises over-acceptance to 0.19 under best-of-8. The exploit also transfers to the second dataset with expert labels: on JudgmentBench grading with the full contract as context, best-of-$N$ selection over author compliance memos drives the single legal judge's acceptance of lawyer-unmet criteria from 0.41 to 0.83 at the Sonnet tier and from 0.33 to 0.51 for the strongest judge tested (Table~\ref{tab:jbattack}; Appendix~B).

The attack and the audit describe one mechanism. Under load a fraction of decisions is settled from surface cues; those decisions default to over-flagging under an unhelpful presentation (the non-adversarial grading bias of the audit), and the adversary's presentation supplies the cue that flips the same decisions to over-acceptance. The overloaded judge under attack simultaneously shows high over-acceptance (above 0.4 on the largest-surface submissions) and reduced over-flagging, confirming that a single unsupported set of decisions is being driven in the adversary's chosen direction.

\section{Sharding Defends the Attack}
\label{sec:defense}

\subsection{The Defense, and Where Gains Are Large}

\begin{figure}[t]
\centering
\includegraphics[width=0.95\textwidth]{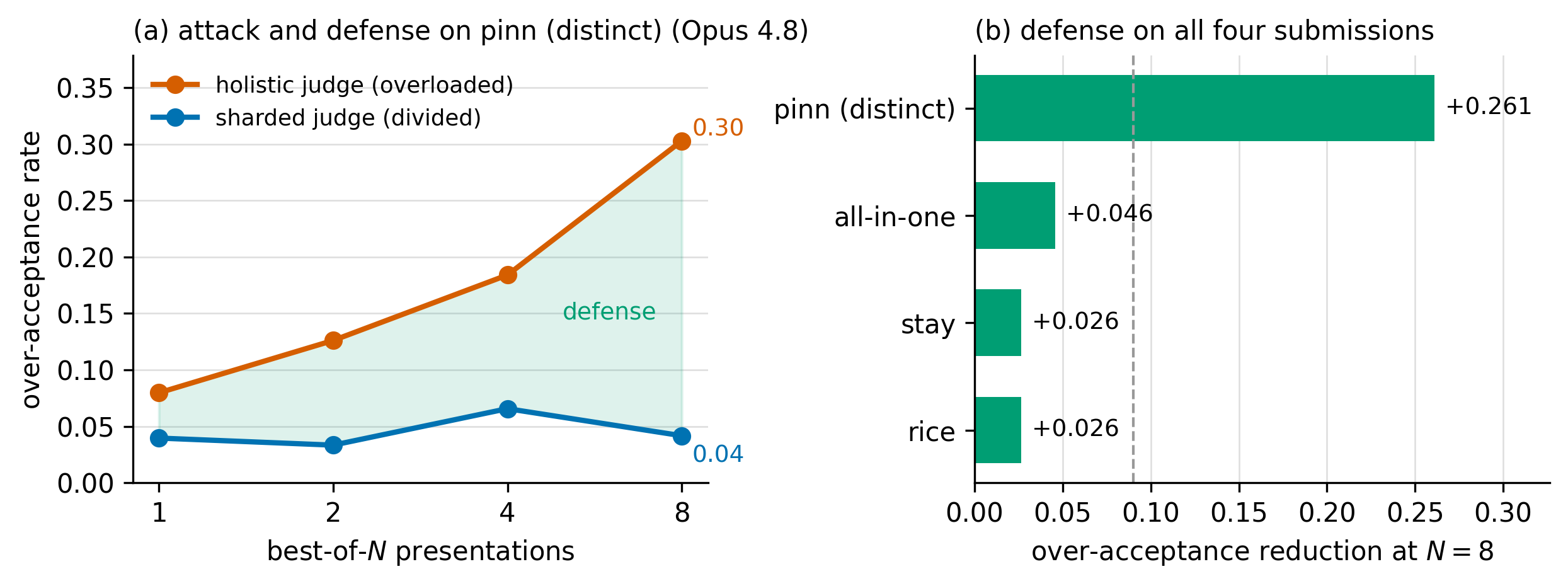}
\caption{Best-of-$N$ presentation attack and its defense on PaperBench grading, judge and adversary Opus 4.8 (all four submissions from one consistent dedup run). (a) On pinn's distinct criterion templates, the attacked holistic judge's over-acceptance of expert-unmet criteria rises with $N$ from 0.08 to 0.30 while the sharded judge holds at 0.04; the shaded region is the defense. (b) The defense at $N=8$ is positive on all four graded submissions, mean 0.090, and largest on the pinn templates (0.261). Sonnet-tier results are in Appendix~B (Table~\ref{tab:attackgrading}).}
\label{fig:attackdef}
\end{figure}

Dividing the decisions removes the unsupported fraction the attack selects, and therefore defends it. On PaperBench grading, sharding the criteria across judge calls reduces the best-of-8 over-acceptance on all four graded submissions at both tiers tested. With judge and adversary Opus 4.8 the cross-submission mean reduction is 0.090 and the largest is 0.261, from 0.303 for the attacked holistic judge to 0.042 for the sharded judge on the pinn templates (Figure~\ref{fig:attackdef}; Table~\ref{tab:opusattack}); at the Sonnet tier the mean is 0.104 (Table~\ref{tab:attackgrading}). As the adversary's search widens, the attacked holistic judge's over-acceptance climbs monotonically while the sharded judge's stays consistently low (Figure~\ref{fig:attackdef}); division removes the unsupported verdicts over which the search selects. On a recoverable submission the defense is two-sided: division reduces both over-acceptance and the miss rate, so it does not trade the exploit for the opposing error (Opus 4.8 on stay: 0.071 to 0.044 over-acceptance and 0.288 to 0.099 miss rate; Sonnet in Table~\ref{tab:twosided}). The defense is the security counterpart of the accuracy result of Section~\ref{sec:accuracy}: the sharded judge, attending to a few criteria per call, verifies them against the submission rather than settling them from the adversary's report.

The defense magnitude across task settings tracks how strongly the decisions contend for shared attention (Figure~\ref{fig:locality}; Table~\ref{tab:codecyber}). On code and cyber verification the attack succeeds (Section~\ref{sec:attack}), but each decision can be evaluated from a local slice of the artifact rather than requiring an integrated model of the whole. The defense is therefore correspondingly small: 0.025 on a concentrated code-verification dataset of interdependent claims (replicating at 0.038 with judge and adversary both Opus 4.8), and 0.022 on the oracle cyber dataset under a length-matched construction. Both effects are positive but smaller than the grading defense of up to 0.261, where the criteria require a shared understanding of a large submission. On the cyber dataset the defense becomes significant, 0.040 with a 95 percent confidence interval of 0.016 to 0.064, when the oracle labels are re-validated by an independent stronger model rather than by the attack judge's own family (Appendix~C). This re-validation both removes a circularity in the label construction and shows that label noise partly diluted the small effect. When the decisions are independent, division does not defend at all: bundling independent code patches into one call and dividing them again changes the cleared-defect rate by negative 0.13 to positive 0.04, within noise of zero, because bundling independent items does not overload a capable verifier and there is no blind spot for division to close. This pattern of a large defense when decisions require integrated context, a small defense when each can be evaluated locally, and no defense when they are independent ties the defense to the same attentional resource as the accuracy result.

\subsection{Adaptive Attacks: When Sharding Alone Is Not Enough}
\label{sec:boundary}

Sharding and opposition repair different missing inputs to a verdict. Under a load-exploiting attack, the judge behaves like a reviewer who leaves part of a long checklist unopened: a favorable cover letter can determine the unchecked boxes. Splitting the checklist into shorter assignments fixes this failure because each box is then opened and compared with the artifact. A persuasion attack survives that repair. It supplies a plausible argument for every criterion, including those the sharded judge now examines carefully. Shortening the checklist cannot supply the counterevidence omitted from that argument; it changes how many cases the judge hears at once, not whether each case is presented from both sides.

\begin{figure}[t]
\centering
\begin{minipage}{0.95\textwidth}
\centering
\begin{subfigure}[t]{0.48\linewidth}
\centering
\includegraphics[width=\linewidth]{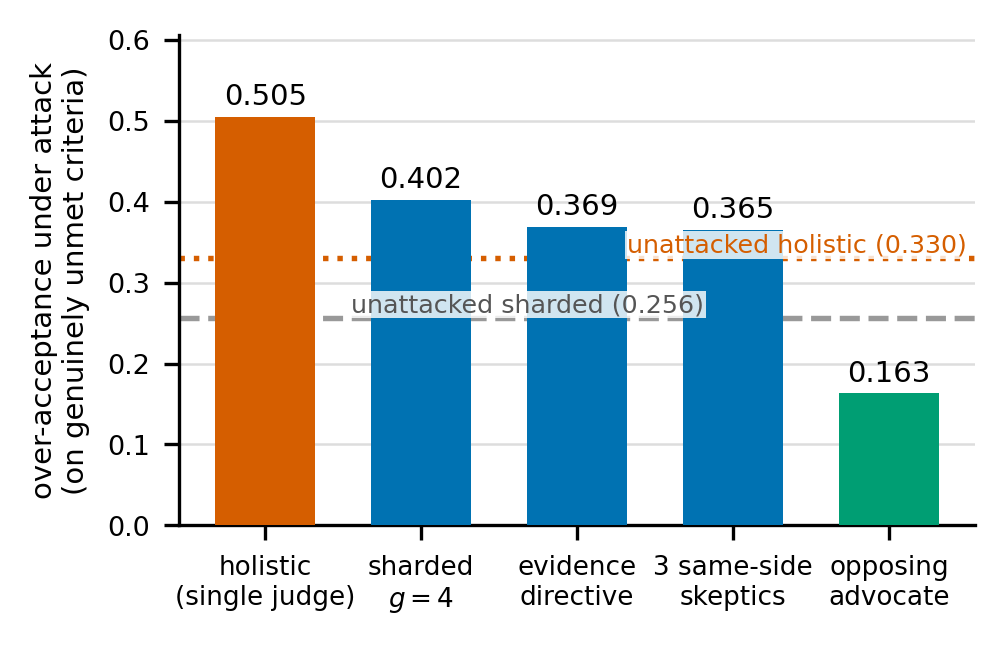}
\caption{Static best-of-6 attack.}
\end{subfigure}\hfill
\begin{subfigure}[t]{0.48\linewidth}
\centering
\includegraphics[width=\linewidth]{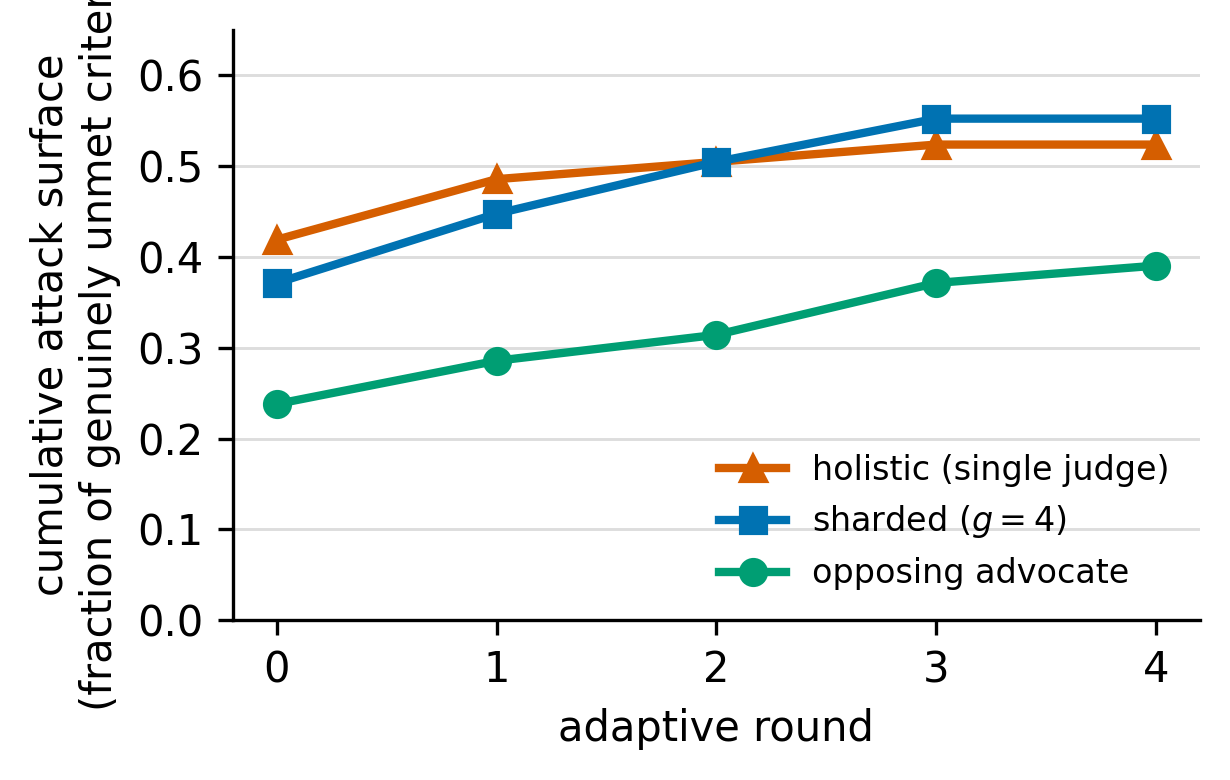}
\caption{Five-round cumulative adaptive attack surface.}
\end{subfigure}
\end{minipage}
\caption{Persuasion attacks and defenses on JudgmentBench with judge and adversary both Opus 4.8. (a) Under a static best-of-6 attack, sharding and two same-side controls reduce over-acceptance only partly, while the opposing advocate reduces it below the unattacked sharded baseline. (b) The cumulative attack surface---the fraction of genuinely unmet criteria accepted at least once by a given round---ends at 0.52 for the single judge, 0.55 for sharding, and 0.39 for the opposing advocate after five rounds of adaptive re-optimization. Per-round realized rates, additional defenses, and false-reject costs are reported in Appendix~B.}
\label{fig:jbattack}
\end{figure}

The two failures call for different interventions. Sharding first ensures that every criterion receives scrutiny. An opposing advocate then ensures that the scrutiny is not one-sided: for each claimed success, a skeptical challenger states the case that the criterion is unmet, and an adjudicator compares the claim with its rebuttal and the underlying evidence. On JudgmentBench with the strongest judge tested, the one-sided compliance memo raises over-acceptance from 0.33 to 0.51. Sharding reduces it to 0.40; an evidence directive and three same-side skeptics reach 0.37, while adding the opposing advocate reduces it to 0.16, below the unattacked sharded baseline of 0.26 (Figure~\ref{fig:jbattack}(a); Table~\ref{tab:jbattackopus}). More attention is therefore sufficient when the attack exploits omitted examination, but balanced evidence is required when the attack persuades a decision that is already being examined.

\emph{We next let the attacker adapt.}
After each round, it learns which criteria the deployed defense still rejects and rewrites the memo to target them. Against sharding alone, each newly targeted criterion still arrives as a one-sided case; across five rounds the attacker obtains acceptance on 0.55 of the genuinely unmet criteria against the strongest judge tested, while opposition generates a fresh counterargument for each target and limits the cumulative rate to 0.39, at a false-reject cost of 0.24 to 0.31 (Figure~\ref{fig:jbattack}(b); Table~\ref{tab:opusadaptive}). The separation is larger at the Sonnet tier: the attacker reaches 0.96 against sharding alone and 0.56 against opposition, which retains kappa 0.40 against the lawyer labels at the final round, compared with 0.12 for the single judge and 0.07 for sharding alone, at a 0.20 false-reject rate on genuinely met criteria (Table~\ref{tab:adaptive}). Sharding is effective against load-based attacks but not universal: it removes decisions settled without examination, while opposition addresses misleading arguments considered during examination.

We have tested this contrast on one dataset, and the result depends on judge capability. On the same legal tasks, the persuasion attack increases over-acceptance less for the strongest judge tested (0.330 to 0.505) than at the Sonnet tier (0.413 to 0.829). The opposing-advocate configuration reduces it to 0.163, below the unattacked sharded baseline, and a judge-by-adversary comparison attributes the greater resistance to the judge rather than to a jointly weaker adversary (Appendix~\ref{app:robustness}).

\section{Within-Shard Structure Is Second-Order; Opposition Defends Adaptive Attacks}
\label{sec:withinshard}
We have shown so far that sharding accounts for the largest first-order benefits.
We next test whether the calls inside a shard should do more than judge independently, for example by arguing opposing sides, reconciling disagreements, cross-examining one another, or decomposing hard criteria into sub-checks. We searched this space systematically, composing opposition (O), reconciliation (R), cross-examination (X), and within-criterion decomposition (amplification, A) on the sharded base across all six grading datasets and four judge tiers, with every arm audited for label leakage (constructions in Appendix~A; full results in Tables~\ref{tab:ladder} to~\ref{tab:attacktypes}).

\begin{figure}[t]
\centering
\includegraphics[width=0.95\textwidth]{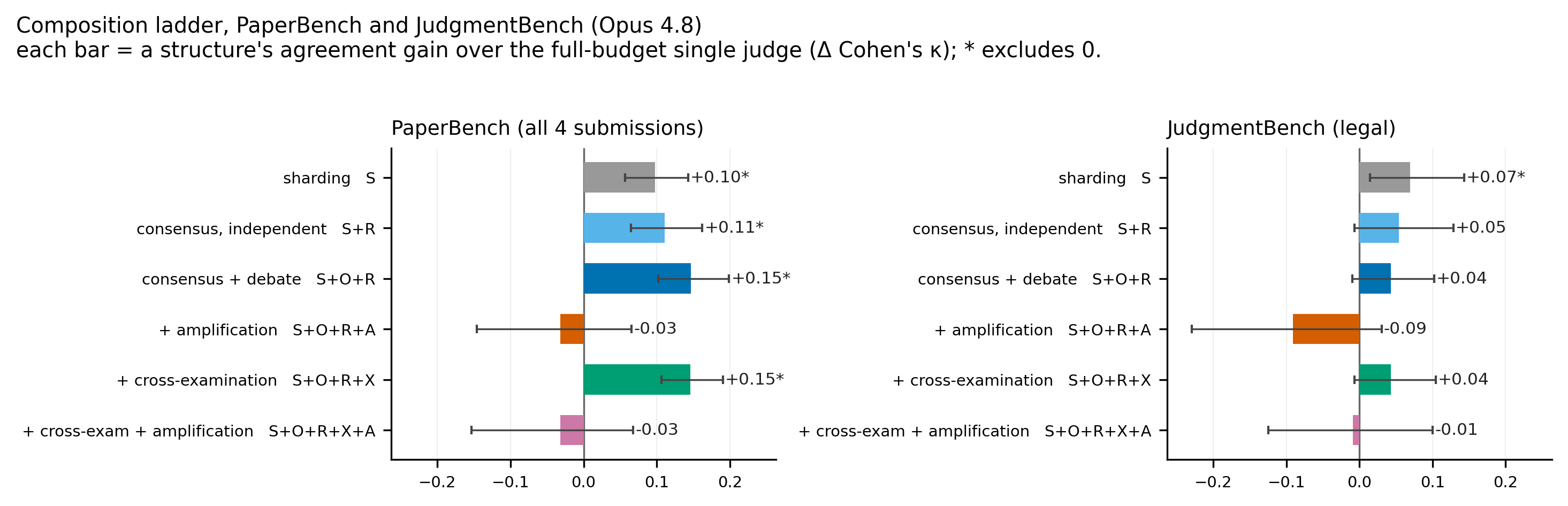}
\caption{Comparison of increasingly elaborate within-shard configurations with the full-budget single judge on the four PaperBench submissions (pooled) and the legal dataset (judge Opus 4.8). Consensus is separated into independent reconciliation (S+R) and reconciliation over opposed sides (S+O+R). Sharding accounts for most of the improvement; consensus adds a small reliable margin attributable to opposition, while configurations that decompose criteria provide no benefit or reduce accuracy. The clinical dataset, where additional budget is more useful than division for a capable judge, is shown separately in Figure~\ref{fig:ladderrob}.}
\label{fig:ladder}
\end{figure}

Every within-shard addition has a smaller effect than division itself (Figure~\ref{fig:ladder}). For the strongest judge tested (Opus 4.8), relative to the full-budget single judge, sharding contributes +0.098 (0.057, 0.143) pooled across the four PaperBench submissions and +0.069 (0.014, 0.143) on the legal dataset. The best addition within a shard, two-sided opposition followed by reconciliation of disagreements (consensus), reaches +0.147 (0.102, 0.199) on the pooled submissions. Cross-examination adds nothing over it (+0.146), both configurations that decompose criteria provide no benefit or reduce accuracy, and no configuration on the legal dataset improves on plain sharding (Table~\ref{tab:ladder2}). At the Sonnet tier the division effect is +0.177 and +0.119 kappa over the single judge on the two primary submissions, whereas consensus adds only +0.040 and +0.030 over plain sharding. An ablation attributes even that smaller margin to targeted reconciliation of disputed decisions rather than to extra passes, prompt diversity, or opposition alone (repeated passes +0.011, diverse prompts +0.009, opposed sides with majority +0.020; Table~\ref{tab:ladder}). Running the full deliberation procedure on decomposed sub-questions adds only +0.002 over decomposition alone at roughly double the cost. Within-criterion decomposition reaches +0.118 on the one dataset whose criteria divide into verifiable sub-checks at the mid-tier judge, but does not generalize: it is negative in most dataset--judge cells, including the legal and clinical datasets, and averages to no benefit or worse across mixed rubrics (Table~\ref{tab:withinbeds}).
Shard-based methods defend against attacks that exploit criteria missed when too many are judged at once, whereas opposition is needed when the attacker persuades the judge criterion by criterion (Table~\ref{tab:attacktypes}; Section~\ref{sec:boundary}). For non-adversarial accuracy, the results favor plain sharding in capacity-sized groups; no within-shard structure substantially improves on it. Reconciled consensus is the only reliably nonnegative addition across datasets (positive or null in ten of eleven cells), but its roughly twofold cost per disputed criterion limits its value, and later deliberation stages are cost-dominated.
Under adaptive attacks, one-sided opposition ($S+O$) keeps the cumulative attack surface lowest: 0.56 at Sonnet and 0.39 at Opus (Figure~\ref{fig:jbattack}(b); Tables~\ref{tab:adaptive} and~\ref{tab:opusadaptive}). Reconciliation reverses this benefit (0.92 and 0.58), while its lower false-reject rate makes the tradeoff explicit.

\section{Interpretation of Why Sharding Works: A Capacity Account, and Its Bounds}
\label{sec:interp}

To explain why sharding works---and when it should not---we interpret the results through a simple capacity account. Each judge run has limited attention that must be divided across its assigned verdicts. Below capacity, the judge can check every decision against the artifact; beyond capacity, it examines only some and resolves the rest from surface cues, defaulting toward rejection when the presentation is unhelpful and toward acceptance when it argues that the work succeeds (Section~\ref{sec:attack}).

Sharding leaves the judge, evidence, and total compute unchanged; it reduces the number of verdicts per run, so the judge can examine every assigned decision. On this account, a stronger judge can cover a larger assignment, so shrinking the assignment gives a weaker judge the per-decision attention of a stronger one, as observed in Section~\ref{sec:accuracy}.

\begin{figure}[t]
\centering
\includegraphics[width=0.95\textwidth]{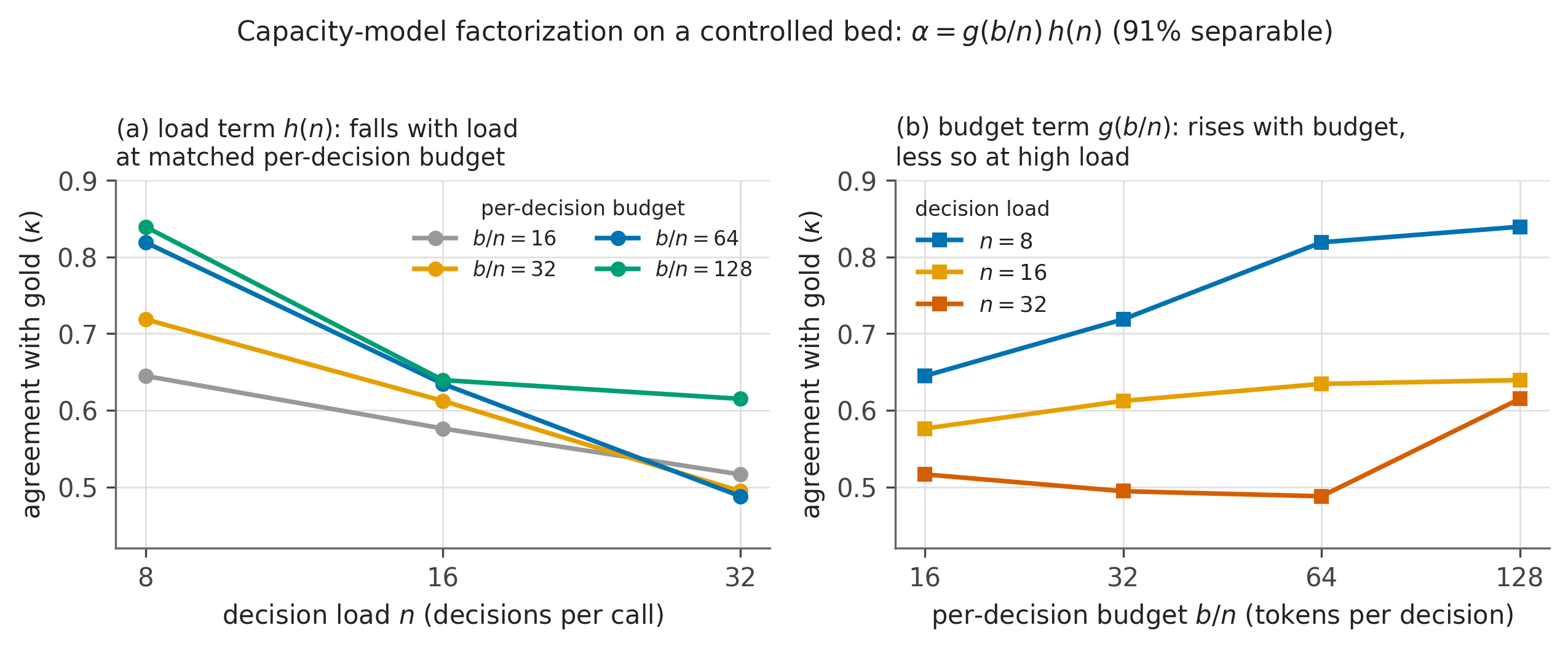}
\caption{Direct test of the capacity account on a controlled synthetic grading dataset. (a) At matched per-decision budget, agreement falls with the decision load, isolating the load term. (b) At fixed load, agreement rises with the per-decision budget, isolating the budget term. A multiplicative fit explains 91 percent of the grid's variance.}
\label{fig:factorization}
\end{figure}

Formally, consider a call evaluating $n$ decisions at budget $b$ and a failed decision $k$ assigned to that call:
$$\Pr[\,s_k \text{ high} \mid y_k=1\,] = \rho_k\,\alpha(n,b),$$
where $\alpha(n,b)=g(b/n)\,h(n)$, $\rho_k$ is recoverability (the probability the failure can be exposed from the evidence at all), $g$ is nondecreasing in the per-decision budget, and $h$ is nonincreasing in the decision load, with $h(n)\approx 1$ below a capacity $n^\star$ and falling past it. On a controlled synthetic dataset the two factors separate cleanly: at matched per-decision budget agreement falls with load, at fixed load it rises with budget, and a multiplicative fit explains 91 percent of the grid's variance (Table~\ref{tab:synth}; Figure~\ref{fig:factorization}). This model links the central results: division raises the load term when budget does not (Section~\ref{sec:accuracy}); the presentation attack selects the unsupported fraction $1-h$, which division removes (Sections~\ref{sec:attack} and~\ref{sec:defense}); and a per-criterion persuasion cue acts outside $\alpha$, which is why division cannot close it (Section~\ref{sec:boundary}). The capability dependence follows directly: $n^\star$ grows with judge capability, so the useful group size shifts with the judge and division has no benefit on workloads a judge can already absorb (Table~\ref{tab:rob}).

\section{Related Work}
\label{sec:related}

Scalable oversight encompasses a range of approaches that go beyond direct evaluation, including debate \citep{irving2018debate,michael2023debate,khan2024debating}, recursive reward modeling \citep{leike2018scalable}, iterated amplification \citep{christiano2018amplification}, prover-verifier games \citep{anil2021pvg,kirchner2024prover}, and weak-to-strong generalization \citep{burns2024weak,lang2026selective}. Recent work has also examined capability gaps and correlated errors among similar models \citep{engels2025scaling,goel2025great}, established complexity-theoretic bounds on agreement guarantees \citep{nayebi2026intrinsic}, and introduced annotated-error datasets for empirical study \citep{recchia2026findtheflaws}. In contrast, we isolate the role of decision load and show that sharding improves both budget-matched single-judge accuracy and robustness.

Model-graded evaluation and LLM-as-a-judge pipelines are now standard \citep{zheng2023judging}, despite documented position, verbosity, self-preference, and confirmation biases \citep{wang2023fair,saito2023verbosity,panickssery2024llm,mahbub2026mitigating,recchia2026confirmation}. Recent benchmarks expose weaknesses on objective and context-dependent comparisons, while theory limits savings over expert labels at the evaluation frontier \citep{tan2025judgebench,xu2025context,dorner2025limits}; robustness varies across prompts and attacks \citep{li2025robustjudge}, and between reranking final answers and guiding intermediate search \citep{zhou2025jetts}. Decision load is distinct because division, rather than prompt debiasing, is the intervention.

Reward overoptimization shows that optimizing against an imperfect proxy can degrade true quality, with best-of-$N$ against a reward model a standard instantiation \citep{gao2023scaling,stiennon2020learning,pan2022effects}. Related work documents attacks on safeguard pipelines \citep{zou2023universal,mckenzie2026stack}, deception \citep{olson2026liecraft}, subtle agent sabotage \citep{kutasov2025shade}, and aggregation-based defenses \citep{cohen2019certified}. Our presentation attack instead targets the oversight model's decision-load blind spot, and we identify sharding as its defense. Long-context evaluation finds degradation when evidence must be inferred rather than matched \citep{modarressi2025nolima}; we hold evidence fixed and vary the number of verdicts. Decomposition has been proposed for grading and relevance judgment \citep{saunders2022self,wu2021recursively,lightman2023verify,george2023factored,farzi2025criteria}; we isolate its accuracy and security benefit against a full-budget control and find within-shard additions second-order in this setting
(Section~\ref{sec:withinshard}).

\section{Discussion}
\label{sec:discussion}

Across three expert-labeled datasets, we show that sharding improves agreement where giving the same budget to one holistic judgment does not. We also show that it removes unsupported verdicts that a presentation adversary can exploit, with the strongest defense when many decisions compete for scrutiny. In these settings, sharding can partly substitute for judge capability. Adaptive persuasion attacks additionally require opposition on the sharded base. Without an adversary, further within-shard communication adds little, so we recommend plain sharding at a calibrated group size whenever decision load reduces accuracy and extra single-call budget fails to recover it.

In Section~\ref{sec:interp}, we explain why increasing compute and dividing work are \emph{not} interchangeable. Increasing $b$ raises the per-decision budget term $g(b/n)$, but the judge must still maintain $n$ distinct verdicts, so the judge's attention shared across $n$ verdicts ($h(n)$) may remain low. Sharding instead lowers $n$ while holding total and per-decision budget fixed. We therefore find that one holistic judgment can spend as much as the divided panel while scrutinizing each decision less. Oversight evaluations should report decision load alongside token or tool budgets.

Our results and analysis show that the optimal group size explicitly depends on three core factors: judge capability, decision difficulty, and the shared understanding required across decisions.
Finally, we use the same account to explain both the security benefit and limits of \emph{pure} sharding. A holistic judge reviewing many criteria at once may still leave some weakly checked, creating an attack surface that an adversary can exploit through presentation alone. Sharding closes this surface by ensuring that each criterion receives sustained scrutiny, but not when an attacker persuades the judge on each criterion individually.
In that adaptive setting, sharding \emph{with} opposition ($S+O$) provides the stronger defense.

\textbf{Limitations.} We test sharding across three expert-labeled domains, including thirteen legal tasks; we test persuasion primarily in legal review where the attack is most natural. The matched-budget advantage is strongest for the most capable judge.
Our result identifies where sharding is most effective: when judges overlook criteria under heavy workloads, rather than when they already handle the workload.
\textbf{AI Use:} GPT assisted with editing; the authors verified all content.

\smallskip
\noindent\textbf{Acknowledgments.} A.N. thanks Geoffrey Irving for early discussions prior to embarking on this work, as well as the Burroughs Wellcome Fund (CASI award), Foresight Institute, and FAR.AI for funding.

\clearpage
\bibliography{\bibliographydatabase}

\clearpage
\setcounter{table}{0}
\setcounter{figure}{0}
\setcounter{algorithm}{0}
\renewcommand{\thetable}{S\arabic{table}}
\renewcommand{\thefigure}{S\arabic{figure}}
\renewcommand{\thealgorithm}{S\arabic{algorithm}}
\appendix

\section{Configurations, Datasets, and Constructions}
\label{app:constructions}

\subsection{The Sharded Arm}
\label{app:shardarm}

The artifact is a set of $K$ independent, self-contained decisions, not a monolithic program. Sharding partitions the decisions and never splits one across calls; every call sees the entire evidence $E$ and differs only in which decisions it returns. The partition is round-robin, decision $k$ to shard $k \bmod S$. The evidence is shared while only the verdicts are divided, and the per-call budget is $B$ (not $B/S$), so the panel spends $SB$ total and $SB/K$ per decision, identical to the full-budget solo control; only the decision load per call differs ($K$ against $K/S$), which is why the contrast (sharded minus full-budget solo) isolates division of work from total compute. In the grading and security datasets the partition is instead into contiguous groups of fixed size $g$, so the number of shards $\lceil K/g \rceil$ grows with load.

\begin{algorithm}[H]
\caption{Sharded oversight}
\label{alg:shard}
\textbf{Input}: decisions and evidence $E$; model $M$ (fixed across arms); per-call budget $B$; shard count $S$\\
\textbf{Output}: scores $s \in [0,1]^K$
\begin{algorithmic}[1]
\STATE partition the decision indices into $S$ disjoint groups $G_1,\ldots,G_S$
\FOR{each $G_j$ \textbf{in parallel}}
\STATE one call of $M$ sees the full evidence $E$, is asked for verdicts on $G_j$ only, at budget $B$, producing $\{\, s_k : k \in G_j \,\}$
\ENDFOR
\STATE \textbf{return} the union of the $S$ disjoint score sets
\end{algorithmic}
\end{algorithm}

For the load sweeps, groups are assigned round-robin to balance the number of decisions; for the grading and security experiments, groups are contiguous and have fixed size $g$. Hidden reference labels are used only after scoring.

\subsection{Replication Details and Core Prompts}
\label{app:replication}

All grading arms use the same neutral judge instruction: ``You are a careful, rigorous grader. You are given a response (or submission) and a list of criteria. For each criterion, decide independently whether the response satisfies it. Judge every criterion on its own merits; do not assume the response is mostly right or mostly wrong.'' The user message contains the full artifact followed by the assigned criteria with neutral identifiers. The required output is \texttt{\{"awards": \{"c0": "yes|no", \ldots\}\}}. Holistic and sharded calls use temperature zero and a 1500-token output cap unless a dataset subsection specifies otherwise. The full-budget call adds: ``Work through every criterion explicitly and one at a time. State the check performed before giving its verdict; do not skim or batch.'' For PaperBench, its output cap is $\min(8000,1500+120K)$; the dataset subsections give the other caps.

The presentation attacker is instructed to write a short, persuasive author note arguing that every criterion is met, citing the strongest available support but not changing the artifact. Legal notes are capped at 350 words; six are sampled at temperature 0.9, and the note maximizing the holistic judge's accept count is selected. The other best-of-$N$ attacks sample eight presentations. The opposing advocate is told to treat the author note as advocacy rather than evidence and to state, for each criterion, the specific missing element, wrong value, unsupported assertion, or other gap. It runs at temperature zero with a 1200-token cap; the adjudicator then receives the artifact, both arguments, and the criteria at temperature zero with a 1600-token cap. In the adaptive attack, the attacker receives only the criteria rejected by the deployed defense and revises the note for five rounds at temperature 0.7.

Hosted inference did not expose a reproducible sampling seed for every model. We therefore report candidate counts and temperatures and retain every generated presentation, selected presentation, verdict, and usage record. Models are pinned to the aliases reported in the paper; where a model does not accept a temperature parameter, the hosted default is used. Structured outputs receive at most two format-correction retries; a still-missing verdict remains missing, and parse rates are checked per arm. Reference labels are never placed in prompts and are joined only during scoring. The code supplement will contain the exact prompt strings, JSON schemas, run commands, final configurations, preprocessing scripts, and per-decision outputs. Inference uses hosted APIs, so accelerator hardware is not controlled by the authors.

\subsection{Synthetic Grading Dataset}

Each item lists $K$ two-digit multiplications, half of which are correct; the incorrect products differ from the true products by small, non-obvious amounts, so the judge must compute them. The dataset isolates an output-space bottleneck. The full-budget arm combines an enlarged token budget with a step-by-step instruction; a budget-only arm (enlarged budget, no instruction) isolates the instruction, whose effect is the difference between the two. The judge is Claude Haiku 4.5.

\subsection{Expert-Graded Replication Dataset (PaperBench)}
\label{app:pb}

JudgeEval of PaperBench: code-replication submissions with hierarchical rubrics whose leaves carry binary expert grades. For the load sweep one submission is fixed as context and $K$ leaf criteria are sampled, six subsets per load; sharded grades one criterion per call ($S=K$) at 1500 tokens, full-budget grades all $K$ in one call at $\min(8000, 1500+120K)$ tokens with the step-by-step instruction; context cap 60000 tokens.

The unit of decision load is one leaf criterion at equal weight, which is the unit a sharded judge grades and is deliberately not PaperBench's native replication score. PaperBench grades all leaves and aggregates bottom-up with per-node weights, so a subtree of near-identical leaves contributes only its parent's weight share and the leaf-count redundancy is neutralized by construction. The flat per-leaf unit does not neutralize it: for a submission with a highly redundant subtree, a random draw of $K$ leaves over-represents one decision. Three of the four submissions have negligible leaf redundancy and are used as released; the pinn submission reduces to 45 distinct criterion templates from 1963 leaves ($\tau \approx 40$, with one template accounting for 77 percent of leaves), so we sample from its distinct templates (dedupe by 60-character prefix) wherever a flat per-leaf number would otherwise conflate repeated leaves with independent decisions. The raw redundant rubric is retained only where the contrast between the two is itself the object of study.

\subsection{Direct-Reading Scope Dataset}

SWE-Bench Pro supplies real multi-file repository tasks graded by hidden tests in prebuilt Docker images; natural pairs combine the correct patch with a failing model-generated patch. The direct load test bundles up to 64 distinct full patches into one read-only context with the per-patch budget set so the bundle fits the window and per-component recoverability is preserved. The verifier is Claude Opus 4.8; we use 8 artifacts per load over 60 pairs and 40 artifacts in the powered execution-mediated run.

\subsection{Oracle Cyber-Verification Dataset}
\label{app:cyber}

Components are confirmed ARVO memory-safety duals (verified vulnerable and fixed source, a reproducing proof-of-concept, and a sanitizer report with an application-frame crash). For each target we reconstruct the call path to the crash from the sanitizer backtrace as a slice of source regions across the involved files, from the fuzzer entry to the crash sink. These regions constitute the artifact context (median 22,000 characters). A claim is a fine-grained technical statement about the bug, labeled true or false against the reconstructed path; the crash site, class, and reachability are the oracle ground truth. Every claim is validated by a full-budget single-claim refuter (Sonnet 4.6, one claim): a claim is retained only if the refuter's verdict equals its intended label, which validates the label in both directions and discards ambiguous claims. Each artifact bundles a balanced set of retained true and false claims for one bug (single-bug concentration). Of 55 targets, 46 are viable at $K=16$ and 45 at $K=12$.

\subsection{Best-of-\texorpdfstring{$N$}{N} Presentation Attack}
\label{app:attack}

The attack holds ground truth fixed and varies only the presentation of the artifact: for grading, an author report asserting every criterion is met; for code, a pull-request description arguing the patch is correct; for cyber, an analyst assessment arguing every claim is valid (adversary Sonnet 4.6). Each of $V=8$ generated presentations is prepended to the fixed artifact and scored; the adversary selects the presentation maximizing accepted decisions. Because the labels are fixed, an accepted genuinely-failed decision is a real over-acceptance. We report the best-of-$N$ over-acceptance (resampled over selection seeds) for the holistic and sharded arms; the defense is holistic minus sharded at $N=8$. The adversary optimizes the judge's own accept-count over all criteria without access to the reference labels; over-acceptance is computed afterward only on criteria labeled unmet, so the adversary never targets those criteria specifically.

\subsection{Label-Surface Controls}

Two constructions prevent an oversight model from separating labels by surface features rather than by verification. Decision identifiers are neutral ($c_0,\dots,c_{K-1}$) assigned after shuffling, so no identifier encodes the label. For the cyber and concentrated-code datasets the false claims are produced by minimally mutating a distinct true claim (same length and structure, wrong only about a derived fact), authored without sight of the label, with a true claim and its own mutation never co-occurring in one artifact; every mutation is refuter-validated false (Appendix~C).

\subsection{Second Grading Dataset with Expert Labels (JudgmentBench)}
\label{app:jb}

JudgmentBench \citep{yang2026judgmentbench} provides expert-lawyer grading of legal work products against professional rubrics (30 tasks, each comprising a prompt, its source documents, and a rubric with per-criterion lawyer grades). We keep the binary rubric items and set the reference label for each (output, criterion) pair to the majority of the annotating lawyers' pass or fail judgments (a criterion passes when awarded points equal the maximum), dropping ties. We prepend the full source contract as shared context rather than the retrieved snippets used by the published autograder; this full-document construction makes each decision require an integrated model of a large artifact, as the load account requires, and matches the conditions under which the lawyers graded. Thirteen tasks have at least twelve binary criteria and a text-extractable contract (median context 34,000 tokens); for one otherwise-qualifying task with scanned exhibits, we transcribe the pages with a multimodal model (Opus 4.8) so the text-only judge can read them. Identifiers are neutral $c_0,\dots,c_{K-1}$. The load curve sweeps criteria per call for a single judge; the arms at full load are holistic, full-budget holistic, and sharded at group size four (Claude Sonnet 4.6), with contrasts reported both over decisions and over the thirteen tasks (a cluster resample, the conservative unit since a task's criteria share one contract). The presentation attack instantiates the author report as a lawyer's compliance memo asserting every criterion is met, generated without the reference labels, six per output, the adversary selecting the memo the deployed holistic judge accepts most. Beyond sharding we evaluate three further defenses on the identical attacked inputs: an evidence-demanding directive on the sharded judge; a pool of three same-side skeptics (the sharded evidence-demanding judge at a temperature spread, per-criterion majority); and an opposing-advocate configuration in which, for each criterion, a skeptical challenger first writes a rebuttal arguing the criterion is unmet and treating the memo as advocacy, and an evidence-demanding adjudicator then decides with that rebuttal in hand. The adaptive variant (Table~\ref{tab:adaptive}) replaces best-of-$N$ sampling with iterative optimization: the attacker submits a memo, reads back which criteria the deployed defense still marks unmet (the per-criterion accept or reject only, never the reference label), and rewrites the memo to target those, for five rounds, run separately against each defense so each attack is tuned to its own defense; a per-round parse-rate check confirms no verdict truncation.

\subsection{Third Grading Dataset with Expert Labels (ROBoto2)}
\label{app:rob}

The dataset contains fully manual expert Cochrane RoB2 risk-of-bias assessments of pediatric randomized-trial reports. Of 620 records, 245 carry a fully manual human assessment and 187 also carry the parsed paper text; these 187 constitute the evaluation dataset. Each trial is assessed on the 22 standardized RoB2 signaling questions across five bias domains, and each question carries the human expert answer, which we collapse to three evaluation classes (yes or probably-yes; no or probably-no; no-information). The signaling answers feed a fixed RoB2 algorithm to a domain-level and overall risk verdict, so faithfulness of the rubric to the overall judgment holds by construction. The questions are integrative rather than extractive: in the released rationales the human assessors cite a single supporting paragraph for only 13 percent of their answers, drawing on the whole study for the remaining 87 percent. The artifact is the paper body (median about 6,000 tokens). The judge answers each signaling question from the report; agreement with the human answer is Cohen's kappa over the three classes. Configurations are holistic (all questions in one call), full-budget holistic (scaled token budget), and sharded at group size $g$, with a group-size sweep $g \in \{1,2,4,8\}$; contrasts bootstrap over trials. Because the 22 questions overload a weaker judge but not a stronger one, results are reported per judge tier on the high-load trials (at least sixteen answered questions, sixty available).

\subsection{Structural Control}
\label{app:controls}

FActScore tests whether the structural profile of the positive results alone produces the load deficit. It contains model-generated person biographies decomposed into atomic facts, each labeled by humans as supported or unsupported against Wikipedia; the irrelevant class is excluded. The judge receives the entity's Wikipedia article as shared context and judges each fact. We use 40 biographies from the generator with the most balanced labels, at ten and thirty facts per call, with Sonnet 4.6 and Haiku 4.5. The configurations are holistic, full-budget holistic, and sharded at group size four; contrasts bootstrap over biographies. Although FActScore has many decisions over shared context, the individual fact checks are easy.

\subsection{Within-Shard Structures}
\label{app:withinshard}

All arms share the sharded base (group size four, full artifact as context, base per-call budget) and the judge model; only the communication among calls varies. Consensus (S+O+R): an advocate call and a skeptic call each return independent verdicts for the shard's criteria (the advocate directed toward definite supported verdicts, the skeptic evidence-demanding); agreements pass through, and each disputed criterion is escalated to a reconciliation call that reads both sides' stated cases and decides. Debate (S+O, one-sided): a lone skeptical challenger writes a rebuttal per criterion and a single adjudicator issues the verdicts with the rebuttal in hand (the Section~\ref{sec:boundary} persuasion defense). Cross-examination (X): each side reads and answers the other's specific claims in one additional round before reconciliation. Amplification (S+A+R): a decomposition call splits a criterion into two to four atomic sub-checks without access to the reference label; each sub-check is judged against the artifact; a recombination call maps the sub-answers to the criterion verdict without re-reading the artifact. The full stack (S+A+O+R+X) runs opposition and cross-examination on every amplified sub-question. Ablation arms isolate the consensus ingredients: repeated same-prompt passes with per-criterion majority (jury), diverse non-opposed prompts with majority, opposed sides with majority and no reconciliation, and independent reconciliation without opposition (S+R, several neutral assessors at a temperature spread, unanimity passing through and disagreements reconciled). For the three-class clinical dataset, the advocate and skeptic argue yes against no, with no-information as the reconciler's fallback. Every arm was audited for label leakage before running: reference labels appear only in scoring, never in any prompt or synthetic sub-question record.
For the JudgmentBench persuasion experiments, the names map to the same primitives: sharded is $S$; the evidence directive is a prompt variant of $S$; three same-side skeptics are an $S{+}R$-like majority-vote arm; the opposing-advocate arm, called ``debate'' in the runner, is one-sided $S{+}O$ and does not include reconciliation. Static best-of-$N$ attacks cover every ladder rung at both judge tiers. Adaptive attacks cover $S$, $S{+}R$, $S{+}O$, and $S{+}O{+}R$ at both tiers; cross-examination, amplification, and the full stack were not run adaptively because they add stages on top of $S{+}O{+}R$, which already performs worse than one-sided opposition under this attack.

\section{Result Tables}
\label{app:tables}

\subsection{Synthetic Grading Dataset and the Factorization}

Synthetic grading dataset (Cohen's kappa; Haiku 4.5): holistic 1.00, 0.56, 0.58, 0.51 at $K=8,16,32,64$; sharded 0.52, 0.69, 0.95, 0.97; full-budget holistic 1.00 at every load. The full-budget single judge recovers completely, so the bottleneck is output space and division is unnecessary; this is the control that distinguishes an output-space bottleneck from a capacity one.

\begin{table}[tb]
\centering
\small
\setlength{\tabcolsep}{3pt}
\begin{tabular}{lcccc}
\toprule
$n \backslash r$ & 16 & 32 & 64 & 128 \\
\midrule
8 & 0.645 & 0.719 & 0.819 & 0.839 \\
16 & 0.576 & 0.612 & 0.634 & 0.639 \\
32 & 0.516 & 0.494 & 0.488 & 0.615 \\
\bottomrule
\end{tabular}
\caption{Capacity-account factorization (synthetic arithmetic; Haiku 4.5; kappa on the planted errors; $n$ facts per call against per-decision token budget $r=b/n$, plain budget with no step-by-step directive; about 550 wrong-fact outcomes per cell, all cells parse-rate 1.00).}
\label{tab:synth}
\end{table}

At matched per-decision budget kappa falls with load (load factors $h$: 1.20, 0.99, 0.84 at $n=8,16,32$); at fixed load it rises with per-decision budget (budget factors $g$: 0.92, 0.96, 1.01, 1.11 at $r=16,32,64,128$). A multiplicative fit $\alpha=g(b/n)\,h(n)$ explains 90.8 percent of the grid variance; the 9.2 percent residual is a sub-multiplicative interaction, budget compensating more within capacity than past it. The $r=8$ column is excluded because the verdict truncated below the parse-rate guard, and $n=4$ because excess budget induces over-reasoning that inflates false alarms. The sweep varies plain token budget; the near-complete recovery of the full-budget arm above comes from its step-by-step directive, an attention intervention, not budget alone.

\subsection{PaperBench Accuracy Tables}

\begin{table}[tb]
\centering
\small
\setlength{\tabcolsep}{3pt}
\begin{tabular}{lcccc}
\toprule
judge & hol. & hol.\ big & shard & shard $-$ big \\
\midrule
Haiku 4.5 & 0.34 & 0.35 & 0.42 & 0.076 (0.035, 0.111) \\
Sonnet 4.6 & 0.59 & 0.65 & 0.75 & 0.096 (0.052, 0.137) \\
Opus 4.6 & 0.68 & 0.75 & 0.79 & 0.040 (0.002, 0.079) \\
\bottomrule
\end{tabular}
\caption{Expert-graded capability sweep (kappa against expert grades; Stay on Topic, 116 leaf criteria; six subsets per load). A four-subset re-draw gives (sharded minus full-budget) 0.087 (0.037, 0.144). On All-in-one (174 leaf criteria, Sonnet 4.6), 0.083 (0.044, 0.117).}
\label{tab:capsweep}
\end{table}

\begin{table}[tb]
\centering
\small
\setlength{\tabcolsep}{3pt}
\begin{tabular}{lcccc}
\toprule
$K$ & hol. & table & shard ($g{=}4$) & shard $-$ table [CI] \\
\midrule
16 & 0.772 & 0.691 & 0.650 & $-$0.040 ($-$0.104, 0.041) \\
32 & 0.729 & 0.672 & 0.807 & 0.135 (0.068, 0.195) \\
64 & 0.586 & 0.689 & 0.753 & 0.064 (0.021, 0.115) \\
116 & 0.553 & 0.661 & 0.737 & 0.076 (0.042, 0.114) \\
\bottomrule
\end{tabular}
\caption{Strongest-monolithic-control test (kappa against expert grades; Stay on Topic; Sonnet 4.6; six subsets per load; every output parsed successfully). The table scaffold is a single full-budget call in which the judge writes one row per criterion, its supporting quote, and a verdict before answering; the forced-citation scaffold (a JudgmentBench-only comparison, kappa 0.568 against sharded 0.611) is a weaker in-call directive omitted here.}
\label{tab:scaffold}
\end{table}

\begin{table}[tb]
\centering
\small
\begin{tabular}{lcccc}
\toprule
$K$ & full & full big & shard4 & shard $-$ full big \\
\midrule
8 & 0.751 & 0.759 & 0.708 & $-$0.049 ($-$0.145, 0.053) \\
24 & 0.800 & 0.805 & 0.793 & $-$0.013 ($-$0.079, 0.044) \\
48 & 0.828 & 0.809 & 0.772 & $-$0.038 ($-$0.070, $-$0.004) \\
64 & 0.833 & 0.812 & 0.757 & $-$0.055 ($-$0.081, $-$0.026) \\
\bottomrule
\end{tabular}
\caption{SWE-Bench Pro direct load test (AUC; Opus 4.8; 4 shards). Single-verifier AUC rises with load (0.751 to 0.833), so the tasks can be evaluated effectively in batches; the execution-enabled run agrees in sign ($-$0.054, standard error 0.037, 40 artifacts). Read-only and execution-enabled verification perform similarly (0.807 and 0.806).}
\label{tab:swebp}
\end{table}

\begin{table}[tb]
\centering
\small
\setlength{\tabcolsep}{3pt}
\begin{tabular}{lcccc}
\toprule
dataset (load) & configuration & miss & over-flag & coverage \\
\midrule
PB stay (116) & holistic & 0.03 & 0.29 & 1.00 \\
PB allinone (174) & holistic & 0.00 & 0.74 & 0.16 \\
PB pinn-45 (45) & holistic & 0.00 & 0.88 & 0.27 \\
\bottomrule
\end{tabular}
\caption{Forced-evidence mechanism audit (error rates at the highest load in each dataset; Sonnet 4.6). miss = P(verdict satisfied $\mid$ true failure); over-flag = P(verdict unsatisfied $\mid$ truly met); coverage = the model's own reported share of decisions it says it checked. Under load the overloaded judge over-flags met criteria rather than missing failures, and coverage falls with group size on the larger rubrics (allinone 1.00 at $g=1$ to 0.16 holistic).}
\label{tab:audit}
\end{table}

Accuracy--cost comparison (Sonnet 4.6; $g$ distinct criteria per call): one decision per call is more expensive and less accurate than another available configuration; on Stay on Topic, $g=32$ reaches kappa 0.860 at 50,524 input tokens against 0.734 at 1,155,404 for $g=1$ and 0.697 for holistic. Isolating the full-budget control, the enlarged token budget alone moves kappa by at most 0.04; the step-by-step instruction is the active component and its sign depends on the dataset.

\subsection{Attack and Defense Tables}

\begin{table}[tb]
\centering
\small
\setlength{\tabcolsep}{3pt}
\begin{tabular}{lcccc}
\toprule
submission & hol.\ $N{=}1$ & hol.\ $N{=}8$ & shard $N{=}8$ & defense \\
\midrule
rice & 0.144 & 0.393 & 0.172 & 0.221 \\
all-in-one & 0.088 & 0.419 & 0.312 & 0.107 \\
pinn (distinct) & 0.023 & 0.154 & 0.103 & 0.051 \\
stay & 0.041 & 0.099 & 0.065 & 0.035 \\
mean & & & & 0.104 \\
\bottomrule
\end{tabular}
\caption{Best-of-$N$ attack and defense, grading (over-acceptance = fraction of expert-unmet criteria accepted; judge and adversary Sonnet 4.6; $V=8$, six criterion-subsets per submission; all four submissions from one consistent run, pinn on distinct templates). Over-acceptance is monotone in $N$ and positive on all four submissions; the defense is positive on all four (mean 0.104), and largest on rice, where the attacked holistic over-acceptance is largest. The per-submission defense is estimated over six criterion-subsets and is correspondingly noisy; the robust claims are its sign on every submission and its ordering, not the third-decimal magnitude.}
\label{tab:attackgrading}
\end{table}

The pinn submission is evaluated on its 45 distinct criterion templates (dedupe by 60-character prefix), not its 1963 redundant leaves, because the load unit is one distinct decision per criterion (Appendix~\ref{app:pb}): under flat per-leaf sampling a random draw of 24 raw pinn leaves contains only about five distinct templates, one repeated roughly eighteen times, which had inflated the raw-rubric defense to 0.192 by scoring near-identical leaves as independent decisions. PaperBench's own weighted-tree aggregation neutralizes this redundancy natively; our per-leaf load unit does not, so we de-duplicate the one redundant submission.

\begin{table}[tb]
\centering
\small
\setlength{\tabcolsep}{3pt}
\begin{tabular}{llcc}
\toprule
submission & quantity & holistic & sharded \\
\midrule
stay (recoverable) & over-acceptance & 0.099 & 0.065 \\
stay & miss (over-flag met) & 0.099 & 0.078 \\
all-in-one (low recov.) & over-acceptance & 0.419 & 0.312 \\
all-in-one & miss & 0.528 & 0.606 \\
\bottomrule
\end{tabular}
\caption{Two-sided property (both-error audit at best-of-8; Sonnet 4.6). On a recoverable submission (stay), division reduces over-acceptance and also reduces the miss rate, so it defends the exploit without trading it for the opposing error. On a low-recoverability submission (all-in-one), division still reduces over-acceptance but raises the miss rate, because when criteria cannot be cleanly verified both arms over-flag and dividing concentrates the over-flagging; pinn behaves the same way (miss 0.779 against 0.556).}
\label{tab:twosided}
\end{table}

\begin{figure}[tb]
\centering
\includegraphics[width=0.95\textwidth]{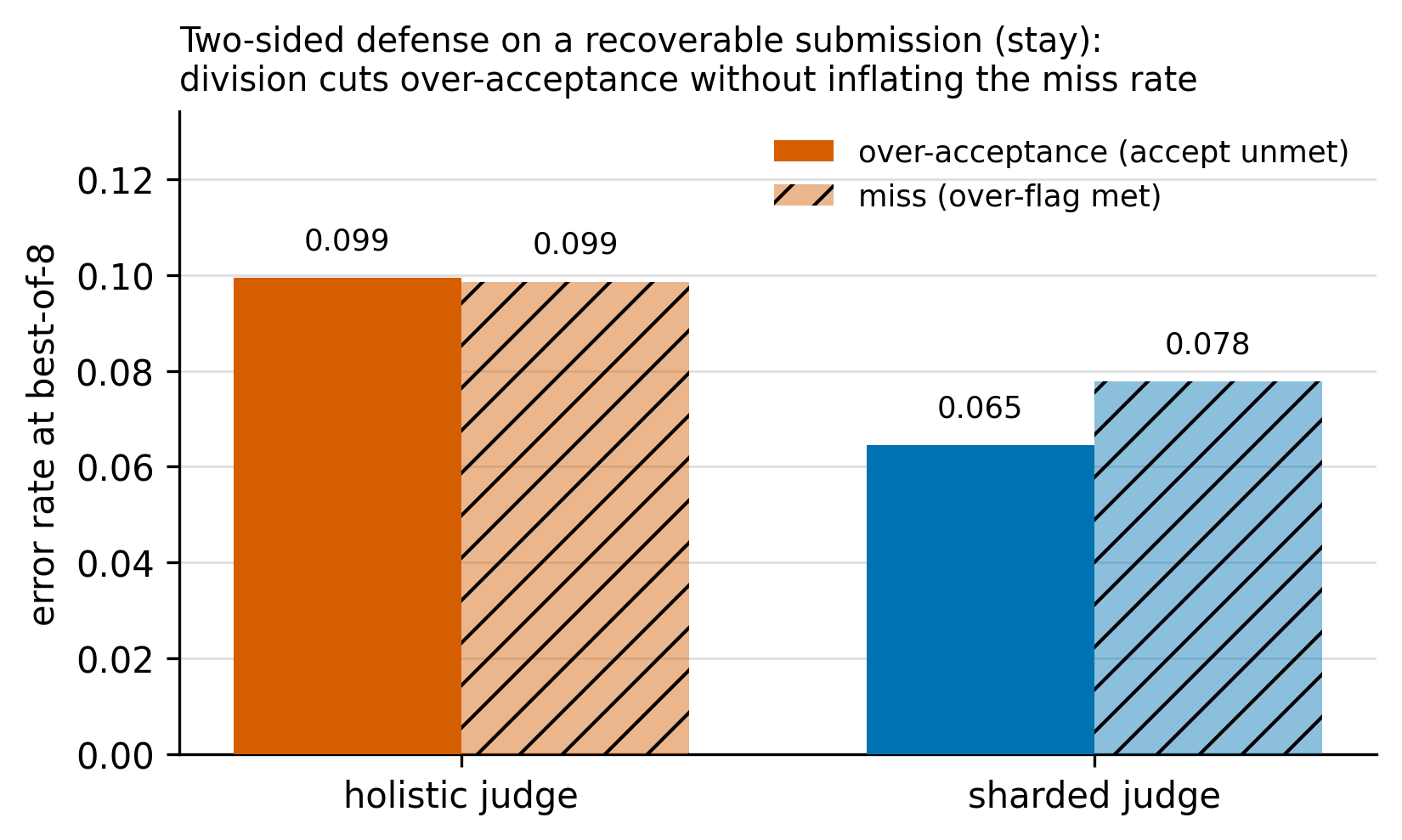}
\caption{Two-sided defense on a recoverable submission (stay) at best-of-8: dividing the criteria reduces the attack's over-acceptance of unmet criteria (0.099 to 0.065) and also reduces the miss rate on met criteria (0.099 to 0.078), so it defends the exploit without trading it for the opposing error.}
\label{fig:twosided}
\end{figure}

\begin{table*}[tb]
\centering
\small
\setlength{\tabcolsep}{2pt}
\begin{adjustbox}{max width=\textwidth}
\begin{tabular}{lcccc}
\toprule
task setting & attack ($N{=}1 \to N{=}8$) & defense $N{=}8$ & 95\% CI & $n$ \\
\midrule
code, independent patches & cleared 0.12 to 0.44; AUC 0.71 to 0.56 & $-$0.13 to $+$0.04 & includes 0 & 8 bundles \\
code, concentrated claims & over-accept to 0.13 & $+$0.025 & -- & 8 \\
cyber, oracle ARVO, $K{=}12$ (same-model ref.) & over-accept 0.111 to 0.189 & $+$0.022 & ($-$0.011, 0.056) & 45 \\
cyber, oracle ARVO, $K{=}12$ (Opus ref.) & over-accept 0.081 to 0.151 & $+$0.040 & (0.016, 0.064) & 42 \\
cyber, oracle ARVO, $K{=}16$ & 0.081 to 0.139 & $+$0.028 & ($-$0.005, 0.065) & 27 \\
\bottomrule
\end{tabular}
\end{adjustbox}
\caption{Attack and defense on code and cyber verification (Sonnet 4.6). The attack succeeds in every task setting; the defense is large only where decisions require shared integrated context, small where they can be evaluated locally, and null where they are independent. The independent-patch null is the strongest evidence for this mechanism: bundling independent patches does not overload a capable verifier (Table~\ref{tab:swebp}), so there is no load-induced blind spot for division to close. On the concentrated interdependent-claims dataset the defense replicates with the strongest model tier tested: with judge and adversary both Opus 4.8 the attacked single verifier reaches 0.150 over-acceptance against the sharded verifier's 0.112, a defense of 0.038 (against 0.025 at the Sonnet tier), still small when decisions can be evaluated locally.}
\label{tab:codecyber}
\end{table*}

\begin{figure}[tb]
\centering
\includegraphics[width=0.95\textwidth]{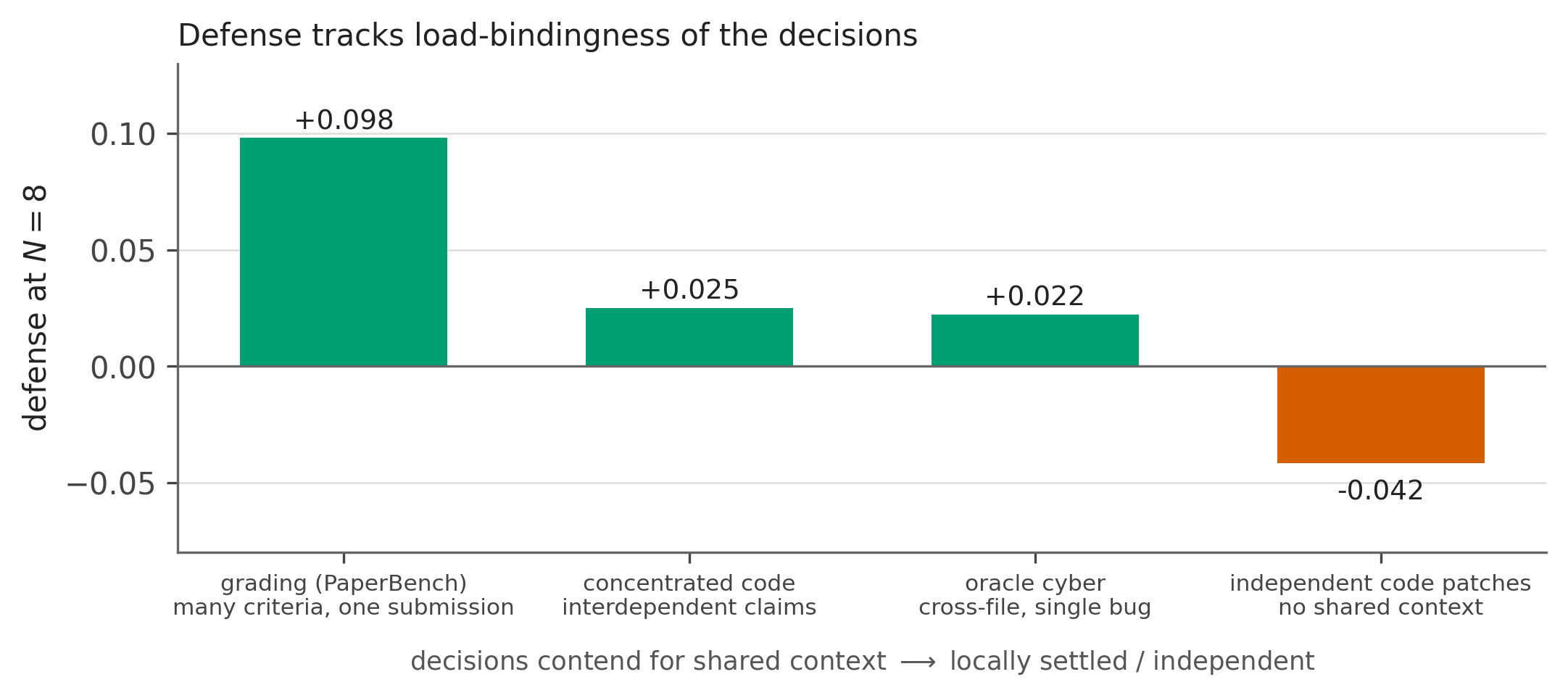}
\caption{The defense at $N=8$ tracks how strongly decision load limits accuracy in each task setting: large on PaperBench grading, where many criteria require one shared understanding of a large submission; small on concentrated code and oracle cyber verification, where each decision can be evaluated from a local slice; and null on independent code patches, where the decisions do not compete for shared attention.}
\label{fig:locality}
\end{figure}

\begin{figure}[tb]
\centering
\includegraphics[width=0.95\textwidth]{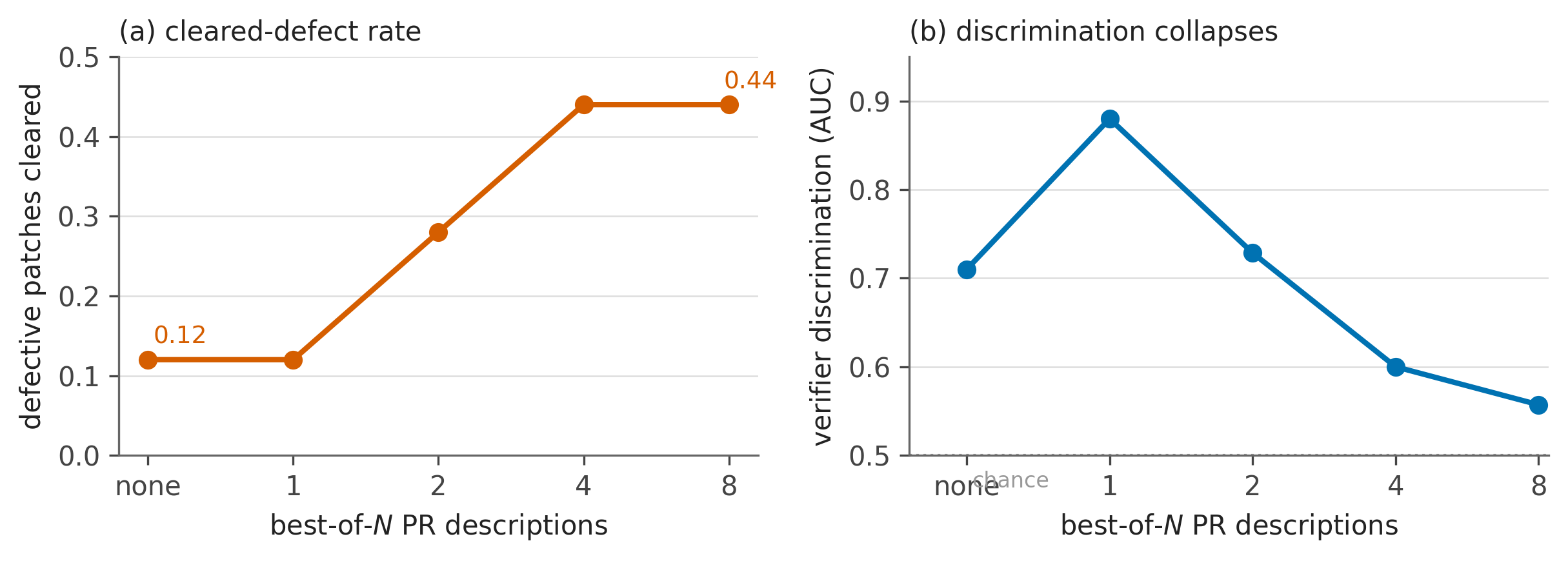}
\caption{Best-of-$N$ presentation attack on a code verifier (independent patches, neutral identifiers). Selecting over $N$ pull-request descriptions raises the fraction of defective patches cleared from 0.12 to 0.44 (a) and lowers the verifier's discrimination from AUC 0.71 to 0.56 (b); a single honest description does not help the adversary, so the exploit is a property of the selection over presentations.}
\label{fig:codeattack}
\end{figure}

\subsection{JudgmentBench Tables}

\begin{table}[tb]
\centering
\small
\setlength{\tabcolsep}{1.5pt}
\begin{tabular}{lcc}
\toprule
arm / contrast & decisions & tasks \\
\midrule
holistic 0.531 & & \\
full-budget 0.572 & & \\
sharded ($g{=}4$) 0.611 & & \\
vs holistic & 0.081 (0.049,\,0.112) & 0.081 (0.009,\,0.148) \\
vs full-budget & 0.039 (0.009,\,0.069) & 0.039 ($-$0.033,\,0.111) \\
\bottomrule
\end{tabular}
\caption{Second dataset with expert labels, accuracy (Cohen's kappa against lawyer labels; JudgmentBench full-document construction; Claude Sonnet 4.6; sharded group size four; 13 legal tasks, 182 outputs). A single judge's agreement falls with load (kappa 0.495, 0.490, 0.309, 0.384, 0.348 at $K=4,8,12,16,20$). The sharded benefit over the single judge excludes zero under both units; over the full-budget control it excludes zero over decisions only and is reported as suggestive. The benefit is positive on 8 of 13 tasks and largest on the highest-load task (183,000-token contract, $+$0.27).}
\label{tab:jbacc}
\end{table}

\begin{table*}[tb]
\centering
\small
\begin{tabular}{lccccc}
\toprule
judge & holistic & full-budget & sharded & sharded $-$ holistic & sharded $-$ full-budget \\
\midrule
Haiku 4.5 & 0.338 & 0.381 & 0.436 & 0.099 (0.014, 0.205) & 0.056 ($-$0.048, 0.157) \\
Sonnet 4.6 & 0.559 & 0.591 & 0.622 & 0.063 ($-$0.003, 0.124) & 0.031 ($-$0.035, 0.090) \\
Opus 4.8 & 0.612 & 0.628 & 0.655 & 0.043 (0.004, 0.091) & 0.027 (0.002, 0.052) \\
\bottomrule
\end{tabular}
\caption{Second dataset with expert labels, capability sweep (kappa against lawyer labels; balanced 13-task, 130-output common subset graded by all three judges; sharded group size four; contrasts task-clustered, the conservative unit). Sharded grading exceeds the single judge at every tier and the benefit shrinks with capability (0.099 to 0.043), as on PaperBench. The benefit persists for the strongest available judge (Opus 4.8, both contrasts excluding zero over the thirteen tasks), because the legal contracts are large (median 34,000, up to 183,000 tokens) and exceed even a strong judge's capacity. The subset is balanced at ten outputs per task, so the Sonnet numbers differ slightly from the fuller sample of Table~\ref{tab:jbacc}.}
\label{tab:jbcap}
\end{table*}

\begin{table}[tb]
\centering
\scriptsize
\setlength{\tabcolsep}{3pt}
\begin{tabular}{lccc}
\toprule
configuration & over-acc. & false-rej. & kappa \\
\midrule
holistic, unattacked & 0.413 & 0.016 & \\
sharded, unattacked & 0.315 & 0.088 & 0.616 \\
holistic, attacked & 0.829 & 0.016 & 0.182 \\
sharded ($g{=}4$), attacked & 0.785 & & \\
shard + evidence directive & 0.703 & 0.036 & 0.295 \\
shard + three skeptics & 0.712 & 0.039 & 0.282 \\
\textbf{opposing-advocate} & \textbf{0.263} & 0.238 & \textbf{0.490} \\
opposed consensus ($S{+}O{+}R$) & 0.751 & 0.032 & 0.249 \\
cross-examination ($S{+}O{+}R{+}X$) & 0.760 & 0.032 & 0.239 \\
amplification ($S{+}A{+}R$) & 0.676 & 0.070 & 0.284 \\
full protocol ($S{+}O{+}R{+}X{+}A$) & 0.676 & 0.075 & 0.277 \\
\bottomrule
\end{tabular}
\caption{Second dataset with expert labels, static persuasion attack and defenses (over-acceptance = fraction of lawyer-unmet criteria accepted; Sonnet 4.6 judge and adversary; best-of-6; 65 outputs, 13 tasks, 438 unmet-criterion decisions). The attack lifts the single judge from 0.413 to 0.829. Division and same-side controls help only partly; one-sided opposition ($S{+}O$) gives the lowest over-acceptance and highest kappa. Adding reconciliation, cross-examination, amplification, or the full protocol does not improve on one-sided opposition.}
\label{tab:jbattack}
\end{table}

\begin{table*}[tb]
\centering
\small
\begin{adjustbox}{max width=\textwidth}
\begin{tabular}{lcccc}
\toprule
defense (under adaptive attack) & per round: round 0 $\to$ final & 5-round cumulative & false-reject & kappa [CI] \\
\midrule
holistic (single judge) & 0.79 $\to$ 0.09 & 0.96 & 0.03 & 0.12 [0.02, 0.25] \\
sharded ($g{=}4$) & 0.75 $\to$ 0.46 & 0.96 & 0.07 & 0.07 [$-$0.02, 0.17] \\
sharded + evidence directive & 0.56 $\to$ 0.35 & 0.93 & 0.04 & 0.34 [0.07, 0.54] \\
sharded + three skeptics ($S{+}R$) & 0.64 $\to$ 0.21 & 0.89 & 0.03 & 0.32 [0.04, 0.54] \\
\textbf{opposing-advocate (debate)} & 0.36 $\to$ \textbf{0.37} & \textbf{0.56} & 0.20 & \textbf{0.40 [0.22, 0.57]} \\
opposed consensus ($S{+}O{+}R$) & 0.64 $\to$ 0.27 & 0.92 & 0.04 & 0.18 [$-$0.02, 0.38] \\
\bottomrule
\end{tabular}
\end{adjustbox}
\caption{Adaptive presentation attack (JudgmentBench; Sonnet judge and adversary; attacker re-optimizes the memo against the deployed judge for five rounds, querying only per-criterion accept or reject; 18 high-load outputs per defense, each with at least five expert-unmet criteria, reproduced on a further 24; every output parsed successfully). Per-round realized over-acceptance is the rate for the single memo submitted in that round. The five-round cumulative surface is the union of unmet criteria accepted in any round, an upper bound that no single memo can realize. Kappa and false-reject are measured at the final round. One-sided opposition uniquely limits the cumulative surface; reconciliation does not.}
\label{tab:adaptive}
\end{table*}

\begin{figure}[tb]
\centering
\includegraphics[width=0.95\textwidth]{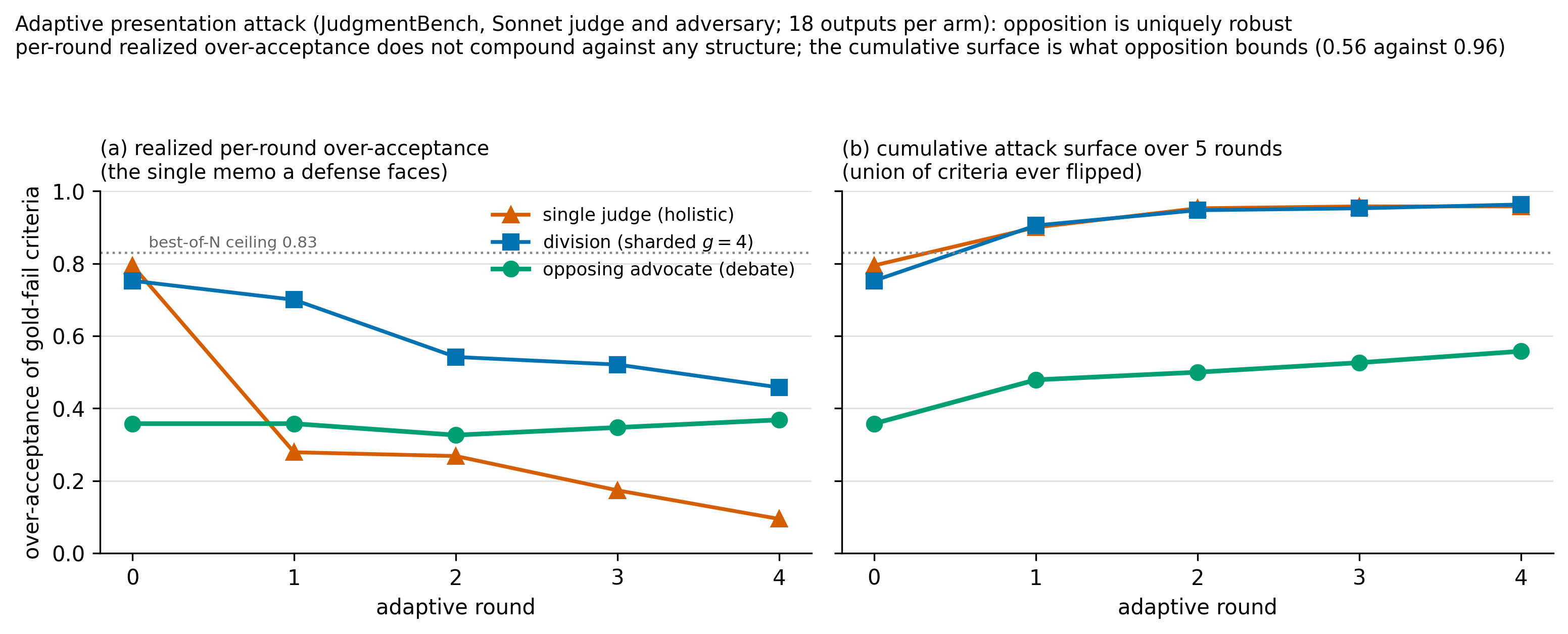}
\caption{Adaptive presentation attack on JudgmentBench: the attacker re-optimizes the memo against the deployed judge each round. The per-round realized over-acceptance of a single memo does not compound against any structure, declining for the single judge and division and staying flat near 0.37 for the opposing-advocate defense; the cumulative attack surface over five rounds (the union of every criterion flipped in any round) is the quantity opposition uniquely bounds, at 0.56 against 0.96 for the others. Bars are bootstrap 95 percent intervals over eighteen high-load outputs.}
\label{fig:adaptive}
\end{figure}

\subsection{ROBoto2 Table}

\begin{table}[tb]
\centering
\small
\setlength{\tabcolsep}{3pt}
\begin{tabular}{lccc}
\toprule
group size $g$ & Haiku 4.5 & (Haiku) shard $-$ hol. & Sonnet 4.6 \\
\midrule
holistic (22) & 0.340 & (baseline) & 0.470 \\
full-budget & 0.340 & 0.000 & 0.456 \\
8 & 0.321 & $-$0.019 ($-$0.068, 0.027) & 0.435 \\
4 & 0.361 & 0.021 ($-$0.027, 0.066) & 0.467 \\
2 & 0.397 & 0.057 (0.005, 0.110) & 0.451 \\
1 & 0.416 & 0.076 (0.033, 0.117) & 0.458 \\
\bottomrule
\end{tabular}
\caption{Third dataset with expert labels, division by group size (kappa against expert RoB2 answers; ROBoto2; high-load trials with at least sixteen answered questions, $n=50$; intervals bootstrap over trials). For the overloaded Haiku judge, agreement rises monotonically as the group shrinks and the contrast excludes zero at $g \le 2$, while the full-budget single judge does not move; for the stronger Sonnet judge, sharding is flat at every group size. The results locate Haiku's capacity for these integrative judgments at about one to two decisions per call.}
\label{tab:rob}
\end{table}

\begin{figure}[tb]
\centering
\includegraphics[width=0.95\textwidth]{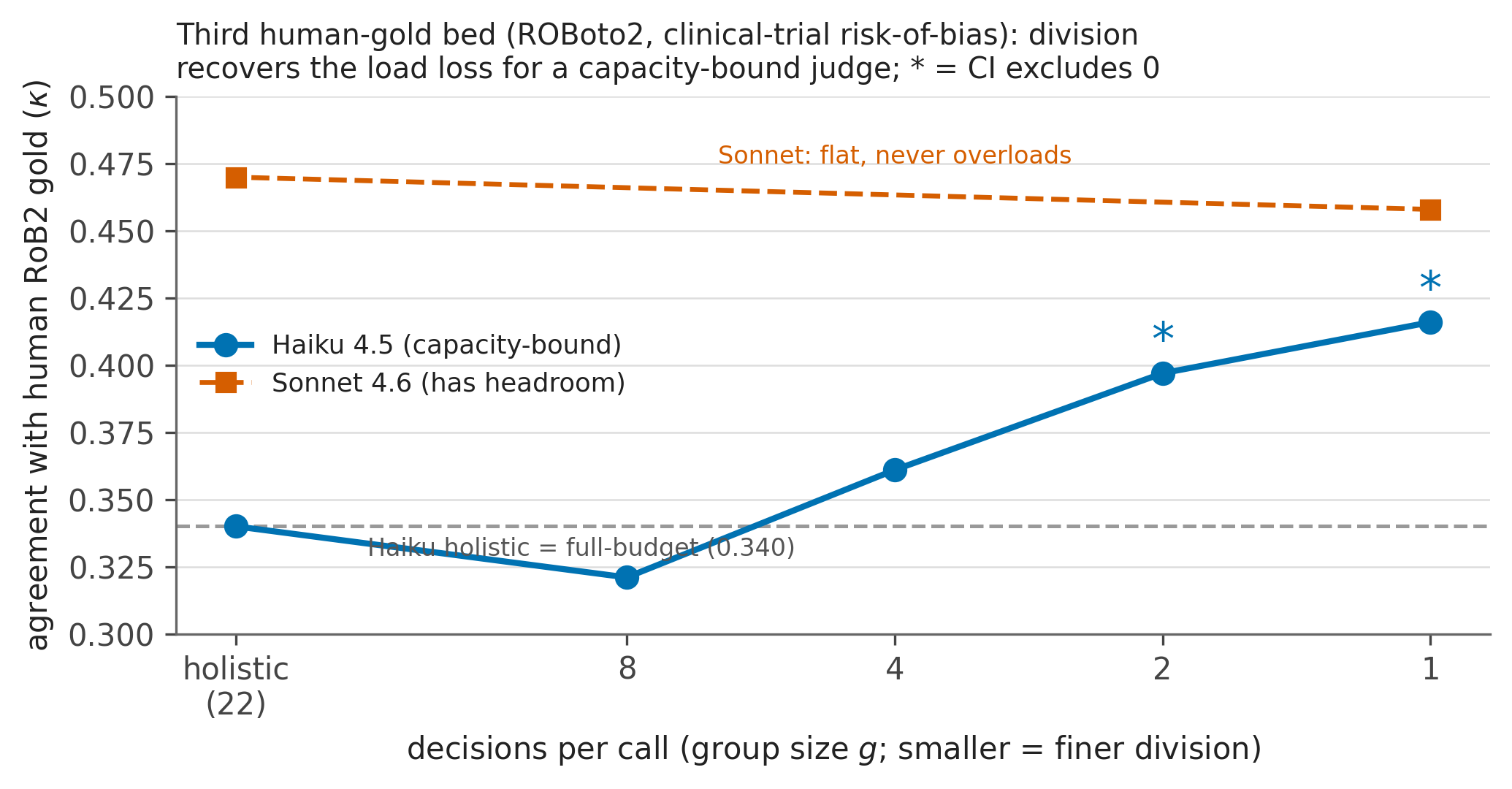}
\caption{Third dataset with expert labels (ROBoto2, clinical-trial risk-of-bias, expert Cochrane RoB2 labels). Agreement with the expert answers against group size on high-load trials. For the overloaded Haiku 4.5 judge, dividing the 22 signaling questions recovers agreement monotonically as the group shrinks, significantly at group sizes of one and two (star = interval excludes zero), while the full-budget single judge does not move; a stronger judge (Sonnet 4.6) is flat.}
\label{fig:rob}
\end{figure}

\subsection{FActScore Structural Control}

\begin{table}[tb]
\centering
\small
\setlength{\tabcolsep}{2pt}
\begin{tabular}{llcccc}
\toprule
judge & load & hol. & big & shard & shard $-$ big \\
\midrule
Sonnet 4.6 & $K{=}10$ & 0.688 & 0.662 & 0.707 & $+$0.045 (0.004, 0.089) \\
Sonnet 4.6 & $K{=}30$ & 0.620 & 0.634 & 0.638 & $+$0.003 ($-$0.022, 0.028) \\
Haiku 4.5 & $K{=}10$ & 0.679 & 0.674 & 0.677 & $+$0.003 ($-$0.037, 0.043) \\
Haiku 4.5 & $K{=}30$ & 0.634 & 0.643 & 0.633 & $-$0.010 ($-$0.036, 0.017) \\
\bottomrule
\end{tabular}
\caption{FActScore structural control (binary kappa against human labels; 40 biographies, ChatGPT generator, positive rate 0.69; sharded at $g=4$; intervals bootstrap over artifacts). Accuracy does not decline with decision load: every configuration sits between kappa 0.62 and 0.71 at every load and judge, and the sharded margin over the full-budget control shrinks with load rather than growing, the reverse of the datasets limited by decision load.}
\label{tab:factscore}
\end{table}

\subsection{Within-Shard Tables}

\begin{table}[tb]
\centering
\small
\setlength{\tabcolsep}{2pt}
\begin{tabular}{lccc}
\toprule
arm & kappa & vs shard (dec.) & vs shard (item) \\
\midrule
holistic & 0.561 & $-$0.177 & $-$0.177 \\
sharded & 0.738 & baseline & baseline \\
jury (repeat, majority) & 0.750 & $+$0.011 & $+$0.011 \\
diverse prompts & 0.747 & $+$0.009 & $+$0.009 \\
single skeptic & 0.724 & $-$0.014 & $-$0.014 \\
opposed sides, majority & 0.759 & $+$0.020 & $+$0.020 \\
one-sided debate & 0.601 & $-$0.137 & $-$0.137 \\
consensus (O$+$R) & 0.779 & $+$0.040 & $+$0.040 \\
consensus $+$ cross-exam & 0.772 & $+$0.034 & $+$0.034 \\
amplification & 0.857 & $+$0.118 & $+$0.118 \\
\bottomrule
\end{tabular}
\caption{Within-shard ablation comparison (kappa against expert grades; primary grading dataset, Sonnet 4.6, eight shuffles; contrast versus plain sharded at group size four). Decision-unit intervals: holistic ($-$0.220, $-$0.136); jury (0.000, 0.024); diverse ($-$0.007, 0.025); skeptic ($-$0.041, 0.014); opposed-majority (0.005, 0.038); one-sided debate ($-$0.178, $-$0.094); consensus (0.022, 0.062); cross-exam (0.013, 0.057); amplification (0.076, 0.163). Item-unit intervals: holistic ($-$0.228, $-$0.118); jury ($-$0.005, 0.036); diverse ($-$0.007, 0.020); skeptic ($-$0.050, 0.027); opposed-majority ($-$0.002, 0.048); debate ($-$0.171, $-$0.105); consensus (0.014, 0.072); cross-exam (0.009, 0.060); amplification (0.080, 0.150). The pattern replicates on the hard second submission (consensus $+$0.030, amplification $+$0.043, debate $-$0.076, jury and diversity null). The full deliberation procedure over amplification is $+$0.002 ($-$0.013, 0.017), showing no additional benefit at roughly double the cost.}
\label{tab:ladder}
\end{table}

\begin{table*}[tb]
\centering
\small
\begin{adjustbox}{max width=\textwidth}
\begin{tabular}{lccc}
\toprule
configuration & PaperBench (pooled) & JudgmentBench & ROBoto2 \\
\midrule
sharding (S) & $+$0.098 (0.057, 0.143) & $+$0.069 (0.014, 0.143) & $-$0.053 ($-$0.096, $-$0.012) \\
independent reconciliation (S$+$R) & $+$0.111 (0.065, 0.162) & $+$0.054 ($-$0.007, 0.128) & $-$0.038 ($-$0.083, 0.004) \\
opposed consensus (S$+$O$+$R) & $+$0.147 (0.102, 0.199) & $+$0.043 ($-$0.009, 0.102) & $-$0.046 ($-$0.094, $-$0.000) \\
plus decomposition (S$+$O$+$R$+$A) & $-$0.032 ($-$0.146, 0.066) & $-$0.091 ($-$0.229, 0.030) & $-$0.080 ($-$0.121, $-$0.035) \\
plus cross-examination (S$+$O$+$R$+$X) & $+$0.146 (0.107, 0.191) & $+$0.043 ($-$0.007, 0.104) & $-$0.045 ($-$0.088, $-$0.003) \\
full stack (S$+$O$+$R$+$X$+$A) & $-$0.032 ($-$0.154, 0.068) & $-$0.009 ($-$0.125, 0.100) & $-$0.084 ($-$0.130, $-$0.036) \\
\bottomrule
\end{tabular}
\end{adjustbox}
\caption{Consensus split into its components (delta versus the full-budget single judge; Opus 4.8; PaperBench pooled across all four submissions, binary kappa; JudgmentBench binary kappa; ROBoto2 three-class kappa; intervals bootstrap over artifacts; S = sharding, R = reconciliation, O = opposition, X = cross-examination, A = within-criterion decomposition). On the pooled PaperBench submissions the step from independent reconciliation to opposed consensus ($+$0.111 to $+$0.147) is larger than the step from sharding to independent reconciliation ($+$0.098 to $+$0.111), so opposition rather than reconciliation alone accounts for the consensus margin. Cross-examination adds nothing over opposed consensus, and both configurations that decompose criteria provide no benefit or reduce accuracy in every dataset group. On the legal dataset no configuration improves on plain sharding, and on the clinical dataset every configuration is at or below the full-budget single judge because additional budget is more useful than division for a capable judge.}
\label{tab:ladder2}
\end{table*}

\begin{figure}[tb]
\centering
\includegraphics[width=0.95\textwidth]{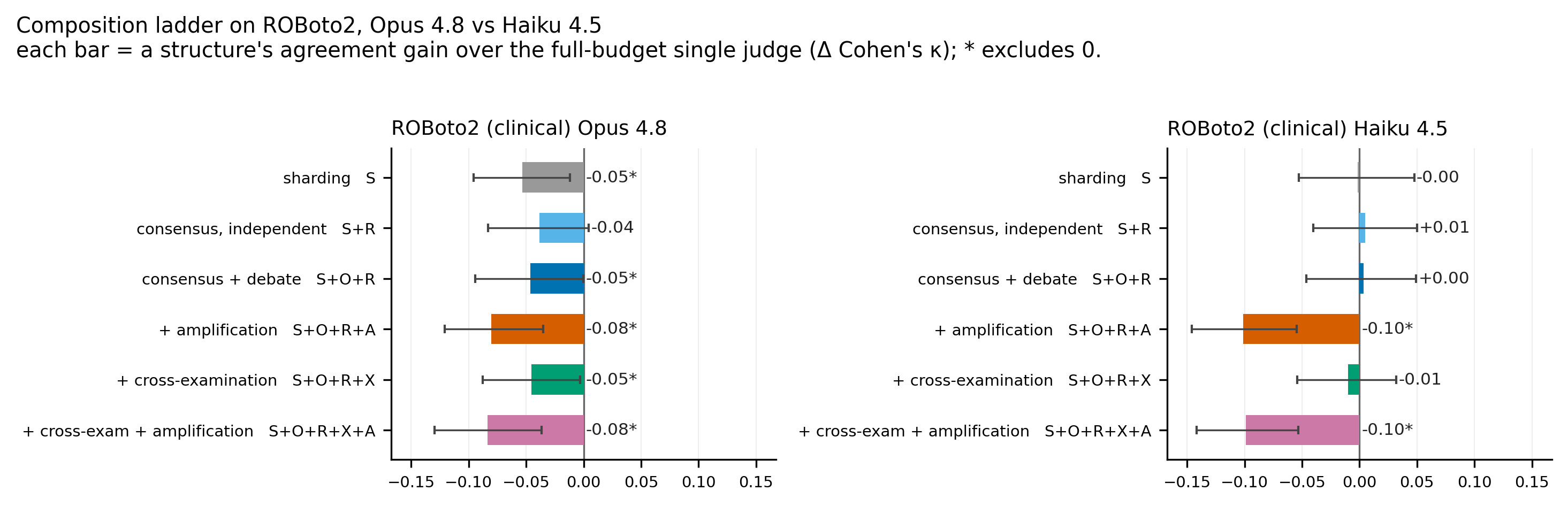}
\caption{Comparison of within-shard configurations on the clinical dataset (ROBoto2), relative to the full-budget single judge, for Opus 4.8 and Haiku 4.5. For the capable judge every configuration is at or below the full-budget single judge because additional budget is more useful than division (Tables~\ref{tab:rob}, \ref{tab:ladder2}). For Haiku the division configurations provide no measurable benefit and the configurations that decompose criteria reduce accuracy, consistent with amplification degrading the calibrated abstention required by the expert labels (Table~\ref{tab:withinbeds}).}
\label{fig:ladderrob}
\end{figure}

\begin{table}[tb]
\centering
\small
\setlength{\tabcolsep}{4pt}
\begin{tabular}{llcc}
\toprule
dataset & judge & consensus & amplif. \\
\midrule
primary & Haiku 4.5 & $-$0.092 & $-$0.043 \\
primary & Sonnet 4.6 & $+$0.040 & $+$0.118 \\
primary & Opus 4.8 & $+$0.012 & $+$0.001 \\
hard submission & Opus 4.8 & $+$0.054 & $+$0.035 \\
legal (non-decomp.) & Sonnet 4.6 & $-$0.021 & $-$0.097 \\
legal & Opus 4.8 & $-$0.027 & $-$0.152 \\
clinical, three-class & Haiku 4.5 & $+$0.005 & $-$0.103 \\
clinical, three-class & Sonnet 4.6 & $+$0.010 & $-$0.174 \\
\bottomrule
\end{tabular}
\caption{Within-shard structures across datasets and judges (contrast versus plain sharded; decision unit; three-class kappa on the clinical dataset). Consensus is positive or within noise of zero in ten of the eleven cells (its one clear negative is the weak judge, $-$0.092); amplification is negative in seven, near zero in one, and positive in three, with its sign depending on whether the criteria decompose into recoverable sub-checks. On the clinical dataset amplification shifts answers away from the no-information class (0.29 to 0.18 for Haiku, 0.19 to 0.14 for Sonnet), degrading the calibrated abstention the expert labels require. Aggregated comparison with the full-budget single judge, pooled over all four submissions: Sonnet consensus $+$0.210 (excluding zero) against amplification $+$0.110 (including zero); Opus consensus $+$0.147 against amplification $-$0.031; for the weak judge every additional configuration moves toward or below zero.}
\label{tab:withinbeds}
\end{table}

\begin{table}[tb]
\centering
\small
\setlength{\tabcolsep}{2pt}
\begin{tabular}{lcc}
\toprule
structure & load-induced attack & persuasion attack \\
\midrule
sharded & $+$0.355 (0.086, 0.582) & $+$0.043 (0.003, 0.084) \\
consensus & $+$0.382 (0.155, 0.592) & $+$0.068 (0.037, 0.103) \\
amplification & $+$0.329 (0.028, 0.595) & $+$0.116 (0.067, 0.163) \\
one-sided debate & $+$0.513 (0.253, 0.741) & $+$0.566 (0.465, 0.669) \\
\bottomrule
\end{tabular}
\caption{Within-shard structures under both attack types (defense = reduction in over-acceptance versus the attacked holistic judge; judge and adversary Sonnet 4.6). Every division-based structure defends the load-induced attack; only opposition fully defends the persuasion attack, while amplification provides a partial defense. Robustness under attack and accuracy without an adversary are separate properties: one-sided debate defends both attacks while harming accuracy in the absence of an adversary (Table~\ref{tab:ladder}).}
\label{tab:attacktypes}
\end{table}

\subsection{Replication with the Strongest Judge Tested (Opus 4.8)}
\label{app:opus}

The main text reports results for the strongest judge tested (Claude Opus 4.8) wherever an Opus 4.8 run of a construction exists; this subsection collects those runs, each holding every other element of its construction fixed. The mid-tier and weaker-tier runs of the same constructions are in the tables and figures above.

\begin{table}[tb]
\centering
\small
\begin{tabular}{lccc}
\toprule
$K$ & holistic & holistic big & sharded \\
\midrule
8 & 0.631 & 0.708 & 0.748 \\
16 & 0.671 & 0.691 & 0.731 \\
32 & 0.569 & 0.636 & 0.757 \\
64 & 0.599 & 0.632 & 0.746 \\
116 & 0.604 & 0.598 & 0.789 \\
\bottomrule
\end{tabular}
\caption{PaperBench load curve on Opus 4.8 (Stay-on-Topic, six subsets per load, sharded at one decision per call; kappa against expert grades). Pooled over loads, (sharded minus holistic) is $+$0.162 (0.135, 0.187) and (sharded minus full-budget) is $+$0.142 (0.107, 0.178), both intervals excluding zero; the sharded margin over the full-budget control at the highest load is $+$0.19. The division effect on Opus 4.8 is larger than the committed Sonnet 4.6 result ($+$0.096), so the main contrast does not diminish with judge generation on this dataset.}
\label{tab:opusload}
\end{table}

\emph{Structured-scaffold control.} On Opus 4.8 the sharded margin over the strongest in-call scaffold grows monotonically with load: holistic, table scaffold, and sharded ($g{=}4$) reach 0.671, 0.671, 0.690 at $K{=}16$; 0.638, 0.729, 0.785 at $K{=}32$; 0.618, 0.651, 0.779 at $K{=}64$; and 0.619, 0.570, 0.743 at $K{=}116$, with (sharded minus table) of $+$0.019 (interval includes zero), $+$0.056, $+$0.129, and $+$0.173 (each excluding zero at $K \ge 32$). At the highest load the table scaffold falls below plain holistic (0.570 against 0.619): the in-call structure itself degrades as the criteria grow, while division scales. Every table output is parsed successfully.

\begin{figure}[tb]
\centering
\includegraphics[width=0.95\textwidth]{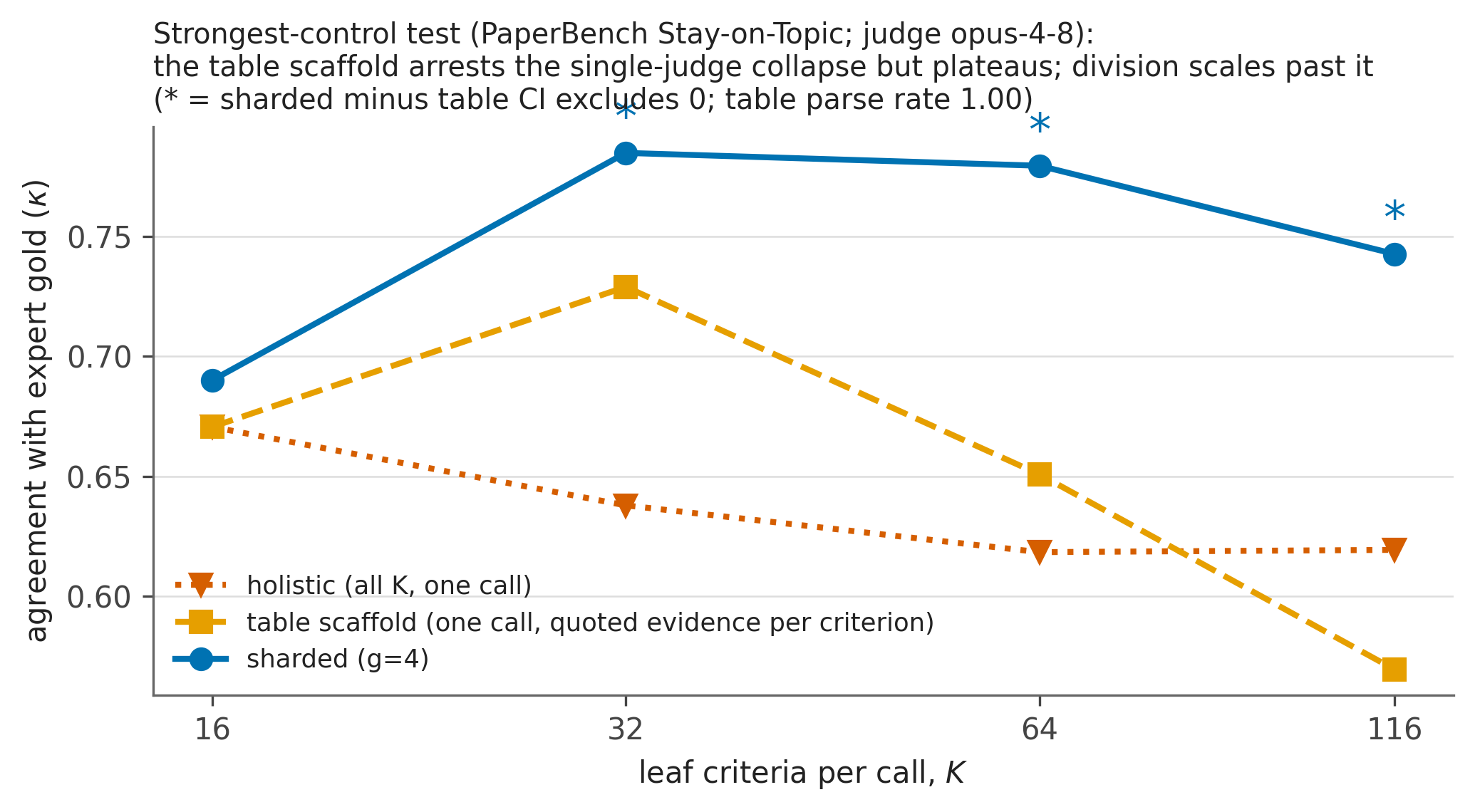}
\caption{Structured-scaffold control on Opus 4.8: the sharded margin over the table scaffold grows monotonically with load, and the scaffold falls below plain holistic at the highest load.}
\label{fig:opusscaffold}
\end{figure}

\emph{JudgmentBench.} The load curve is 0.672, 0.656, 0.685, 0.625, 0.538 at $K = 4, 8, 12, 16, 20$, so the decline begins near twelve criteria for Sonnet 4.6 but near twenty for Opus 4.8, showing within one dataset how the effect changes with model capability. The full-load configurations reproduce Table~\ref{tab:jbcap}: holistic 0.612 (0.473, 0.712), full-budget 0.628 (0.489, 0.724), sharded 0.655 (0.513, 0.743).

\emph{ROBoto2.} Sharded three-class kappa is 0.395, 0.422, 0.403, 0.372 at $g = 1, 2, 4, 8$ against holistic 0.405, flat at every group size. The 22 questions remain within Opus 4.8's capacity, unlike the capacity-limited Haiku results (Table~\ref{tab:rob}; Figure~\ref{fig:rob}).

\begin{table}[tb]
\centering
\small
\setlength{\tabcolsep}{3pt}
\begin{tabular}{lcccc}
\toprule
submission & hol.\ $N{=}1$ & hol.\ $N{=}8$ & shard $N{=}8$ & defense \\
\midrule
pinn (distinct) & 0.080 & 0.303 & 0.042 & 0.261 \\
all-in-one & 0.152 & 0.408 & 0.362 & 0.046 \\
rice & 0.166 & 0.253 & 0.227 & 0.026 \\
stay & 0.034 & 0.071 & 0.044 & 0.026 \\
mean & & & & 0.090 \\
\bottomrule
\end{tabular}
\caption{Load-induced attack and defense with judge and adversary both Opus 4.8 ($K{=}24$, $V{=}8$, six subsets per submission, all four from one consistent dedup run; over-acceptance of expert-unmet criteria). The attack succeeds against the strongest judge tested (holistic over-acceptance at least doubles under best-of-8 on every submission) and division defends on all four, mean 0.090, comparable to the Sonnet 4.6 mean of 0.104. The two-sided property also replicates on a recoverable submission (stay): division reduces over-acceptance (0.071 to 0.044) and the miss rate together (0.288 to 0.099).}
\label{tab:opusattack}
\end{table}

\begin{figure}[tb]
\centering
\includegraphics[width=0.95\textwidth]{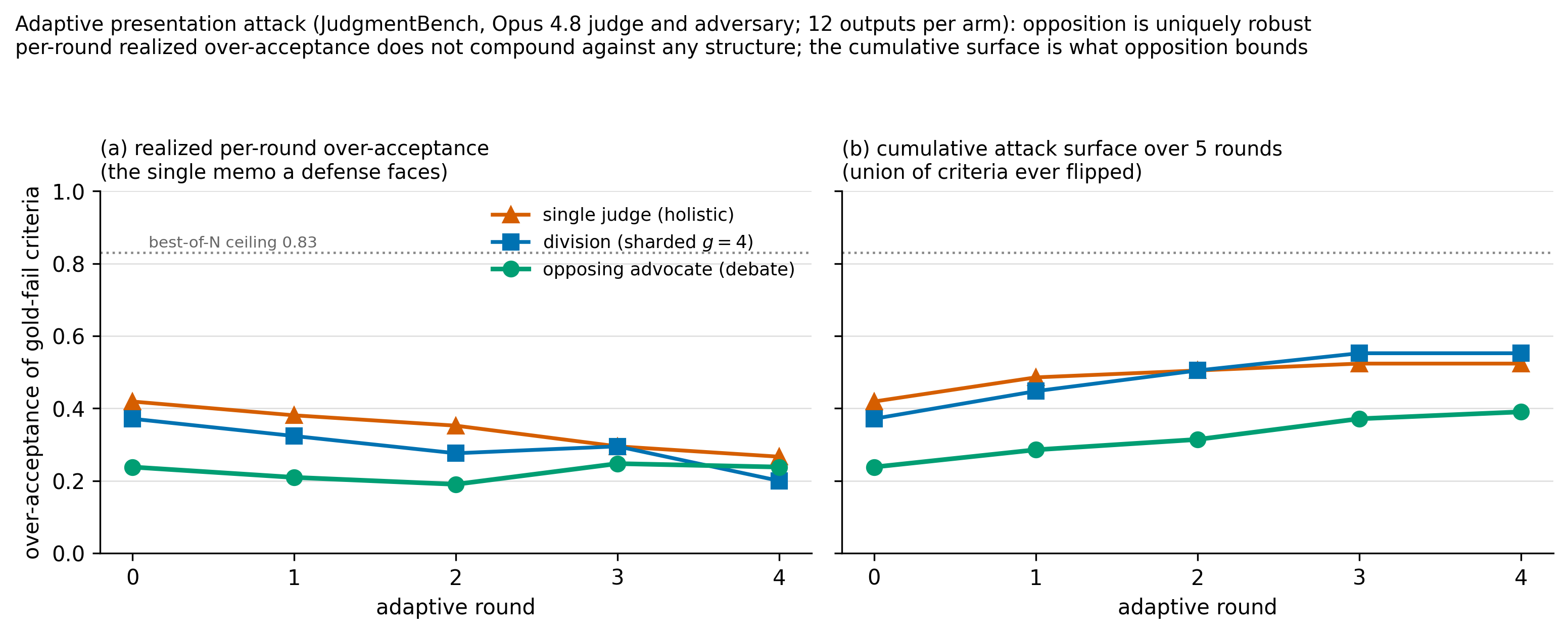}
\caption{Adaptive presentation attack on JudgmentBench with judge and adversary both Opus 4.8: per-round over-acceptance does not grow against any configuration, while opposition has the smallest five-round cumulative surface (0.39 versus 0.47--0.58 for the other defended arms).}
\label{fig:opadaptive}
\end{figure}

\emph{Persuasion attack, defenses, and the adaptive variant} (78 attacked outputs for the configuration comparison, twelve high-load outputs per configuration for the adaptive variant, with every output parsed successfully at every round): unattacked over-acceptance is 0.330 (holistic) and 0.256 (sharded); the best-of-6 memo lifts the holistic judge to 0.505, compared with the Sonnet increase from 0.413 to 0.829, so the persuasion effect is much smaller for the strongest model tier tested. Under attack, division reaches 0.402, the evidence directive 0.369, three same-side skeptics 0.365, and one-sided opposition 0.163, below the unattacked sharded baseline. The full static ladder is in Table~\ref{tab:jbattackopus}: adding reconciliation, cross-examination, amplification, or all stages does not improve on one-sided opposition. The adaptive attack does not grow across rounds against either judge. For Opus, the per-round rate declines for the holistic and sharded defenses and stays near 0.24 for opposition; the added same-side and consensus rows are reported in Table~\ref{tab:opusadaptive}. The cumulative surface reaches 0.52 for the single judge, 0.55 for division, 0.47 for same-side skeptics, and 0.58 for opposed consensus, but only 0.39 for opposition (Table~\ref{tab:opusadaptive}; Figure~\ref{fig:opadaptive}). The opposing advocate retains its false-reject cost (0.24 to 0.31). The stronger judge starts lower, consistent with its greater resistance to best-of-$N$ selection, but the caveat that the persuasion results come from one dataset applies here equally.

\begin{table*}[tb]
\centering
\small
\begin{adjustbox}{max width=\textwidth}
\begin{tabular}{lcccc}
\toprule
defense (under adaptive attack) & per round: round 0 $\to$ final & 5-round cumulative & false-reject & kappa [CI] \\
\midrule
holistic (single judge) & 0.42 $\to$ 0.27 & 0.52 & 0.12 & 0.47 [0.19, 0.67] \\
sharded ($g{=}4$) & 0.37 $\to$ 0.20 & 0.55 & 0.14 & 0.53 [0.27, 0.72] \\
sharded + evidence directive & 0.31 $\to$ 0.26 & 0.52 & 0.13 & 0.55 [0.31, 0.72] \\
sharded + three skeptics ($S{+}R$) & 0.30 $\to$ 0.28 & 0.47 & 0.14 & 0.52 [0.29, 0.68] \\
\textbf{opposing-advocate ($S{+}O$)} & 0.24 $\to$ 0.24 & \textbf{0.39} & 0.28 & 0.46 [0.23, 0.70] \\
opposed consensus ($S{+}O{+}R$) & 0.35 $\to$ 0.30 & 0.58 & 0.12 & 0.46 [0.25, 0.61] \\
\bottomrule
\end{tabular}
\end{adjustbox}
\caption{Adaptive presentation attack with judge and adversary both Opus 4.8 (JudgmentBench; the Table~\ref{tab:adaptive} construction with twelve high-load outputs per configuration; every output parsed successfully at every round). The per-round realized rate is the over-acceptance of the single memo submitted in a round; the five-round cumulative value is the union of accepted unmet criteria across rounds. The attack does not grow across rounds against any configuration. Opposition uniquely limits the cumulative surface (0.39, versus 0.47--0.58 for the other defended arms), but it also has the highest false-reject rate.}
\label{tab:opusadaptive}
\end{table*}

\begin{table*}[tb]
\centering
\small
\begin{tabular}{lcccc}
\toprule
configuration & over-acceptance & false-reject & kappa & primitives \\
\midrule
holistic, unattacked & 0.330 & -- & -- & -- \\
sharded, unattacked & 0.256 & -- & -- & $S$ \\
holistic, attacked & 0.505 & 0.038 & 0.498 & -- \\
sharded, attacked & 0.402 & 0.057 & 0.574 & $S$ \\
evidence directive & 0.369 & 0.069 & 0.590 & $S$ + directive \\
three same-side skeptics & 0.365 & 0.065 & 0.598 & $S{+}R$-like \\
\textbf{opposing advocate} & \textbf{0.163} & 0.262 & 0.551 & $S{+}O$ \\
opposed consensus & 0.409 & 0.061 & 0.563 & $S{+}O{+}R$ \\
cross-examination & 0.377 & 0.054 & 0.601 & $S{+}O{+}R{+}X$ \\
amplification & 0.337 & 0.162 & 0.509 & $S{+}A{+}R$ \\
full protocol & 0.340 & 0.146 & 0.524 & $S{+}O{+}R{+}X{+}A$ \\
\bottomrule
\end{tabular}
\caption{Static persuasion attack and defenses on JudgmentBench with judge and adversary both Opus 4.8 (best-of-6; 78 attacked outputs). One-sided opposition ($S{+}O$) is the only arm that brings over-acceptance below the unattacked sharded baseline of 0.256. The higher rungs, including opposed consensus, do not improve robustness under this attack even though consensus helps accuracy without an adversary.}
\label{tab:jbattackopus}
\end{table*}

\subsection{Additional Replications}
\label{app:robustness}

\emph{Second submission for the load sweep} (the primary sweep repeated on All-in-one, Sonnet 4.6, six subsets per load; sharded at one decision per call; intervals bootstrap over subsets): holistic 0.065, 0.120, 0.166, 0.060, 0.046 and full-budget 0.137, 0.166, 0.211, 0.096, 0.025 and sharded 0.137, 0.166, 0.166, 0.177, 0.165 at $K = 8, 16, 32, 64, 174$. The sharded margin over the single judge is $+$0.070 ($-$0.065, 0.193), $+$0.044 ($-$0.061, 0.141), $-$0.000 ($-$0.031, 0.035), $+$0.118 (0.084, 0.151), and $+$0.140 (0.091, 0.145), and over the full-budget control $+$0.000, $+$0.000, $-$0.045, $+$0.082, $+$0.140. The division result is not specific to one document: on a second submission's full-load sweep the sharded margin over both controls grows with load, reaching $+$0.140 at the full 174 criteria, both intervals excluding zero. The low-load and $K{=}32$ cells are within noise (this submission has low absolute agreement, 0.05 to 0.21, so the per-cell intervals are wide), and the effect is carried by the high-load cells where decision load limits accuracy, as on Stay-on-Topic.

\emph{Judge-by-adversary comparison} (JudgmentBench persuasion attack, over-acceptance of lawyer-unmet criteria under the attacked holistic judge, all four judge-adversary pairs; the diagonal cells are the main results): judge Sonnet 4.6 reaches 0.829 against adversary Sonnet 4.6 and 0.716 against adversary Opus 4.8; judge Opus 4.8 reaches 0.616 and 0.505. Holding the adversary fixed, upgrading the judge lowers over-acceptance (Sonnet adversary: 0.829 to 0.616; Opus adversary: 0.716 to 0.505), so the strongest judge's greater resistance to the persuasion attack is a property of the judge and not an artifact of a jointly weaker adversary: the same Sonnet adversary that drives the Sonnet judge to 0.829 reaches only 0.616 against the Opus judge. Holding the judge fixed, upgrading the adversary does not strengthen the attack (it lowers over-acceptance slightly), so adversary capability does not limit the attack in the tested range. The sharded configuration shows the same pattern (0.491 for judge Opus against the Sonnet adversary, 0.586 for judge Sonnet against the Opus adversary).

\section{Label-Surface Controls and Critical Decisions}
\label{app:controlsC}

\textbf{Surface-feature ablation.} For a generated claim dataset to test verification rather than surface pattern-matching, a text-only classifier must not separate the labels. Under naive authoring (true and false claims generated together) a length-only classifier separates the two classes at AUC 0.70 to 0.76. Under the length-matched blind construction it falls to 0.545, near chance, and all reported cyber and concentrated-code numbers use it. The defense point estimate is preserved under the control (0.028 length-matched against 0.032 before it at $K=16$), so the effect is not a surface artifact; removing the tell raises the unattacked over-acceptance from 0.025 to about 0.09, showing that the surface feature had assisted detection in both configurations.

\textbf{Independent-refuter validation of the oracle labels.} The cyber labels are validated by a full-budget single-claim refuter, and the attack judge is from the same model family, which could make the reference labels circular. Re-validating every claim with an independent stronger model (Opus 4.8) and keeping only the claims it confirms, Opus agrees with 88 percent of the true labels and 94 percent of the false labels, so the labels are largely model-independent. Re-running the attack on the Opus-validated dataset with the same Sonnet judge gives a defense of 0.040 with a 95 percent confidence interval of 0.016 to 0.064, which excludes zero, against 0.022 (interval including zero) for the same-model dataset; the stronger refuter's cleaner labels sharpen the signal, and the effect is not an artifact of same-model validation.

\textbf{Critical statistical and modeling decisions.} Every contrast is a bootstrap 95 percent confidence interval over items (an artifact in the code datasets, a graded submission in the grading datasets); any sharding contrast on fewer than thirty coding artifacts is treated as underpowered and reported with its per-artifact gap standard error, a threshold calibrated to the 0.36 per-artifact dispersion of the coding-task datasets and not applied to near-degenerate cells. Oracle cyber labels are retained only when a full-budget single-claim refuter agrees with the intended label in both directions, so an uncaught defect under load is a load effect and not a label error. Judges and verifiers are pinned Claude-family models (Haiku 4.5, Sonnet 4.6, Opus 4.6, Opus 4.8).

\end{document}